\documentclass[10pt,a4paper]{article}
\usepackage{graphicx,amsfonts,amssymb,amsmath, hyperref, setspace,wasysym}
\usepackage{a4wide}
\usepackage{authblk,enumerate,multirow}
\usepackage[normalem]{ulem}
\newcommand{\stkout}[1]{\ifmmode\text{\sout{\ensuremath{#1}}}\else\sout{#1}\fi}

\usepackage{caption}
\newif\ifhyper
\hypertrue
\ifhyper
\hypersetup{
   citecolor = {green},
   colorlinks = {true}, % false
   linkcolor = {black},
   urlcolor = {blue} % magenta
}
\fi
\newcommand{\beq}{\begin{equation}}
\newcommand{\eeq}{\end{equation}}
\newcommand{\beqa}{\begin{eqnarray}}
\newcommand{\eeqa}{\end{eqnarray}}
\newcommand{\ket} [1] {\vert #1 \rangle}
\newcommand{\bra} [1] {\langle #1 \vert}
\newcommand{\braket}[2]{\langle #1 | #2 \rangle}

\newcommand{\tr}{\mathop{\mathrm{tr}}}

\def\bra#1{\langle#1\vert}
\def\ket#1{\vert#1\rangle}

\def\Longarrow{\protect\@lra}
\def\@lra{\relbar\joinrel\relbar\joinrel\relbar\joinrel%
          \relbar\joinrel\rightarrow}

\begin{document}

\title{Mathematical foundations of matrix syntax}

\author[1,2,3]{Rom\'an Or\'us} 
\affil[1]{Donostia International Physics Center, Paseo Manuel de Lardizabal 4, E-20018 San Sebasti\'an, Spain}
\affil[2]{Ikerbasque Foundation for Science, Maria Diaz de Haro 3, E-48013 Bilbao, Spain}
\affil[3]{Institute of Physics, Johannes Gutenberg University, 55099 Mainz, Germany}

\author[4]{Roger Martin} 
\affil[4]{Faculty of Environment and Information Sciences,  Yokohama National University, 79-7, Tokiwadai, Hodogaya-ku, Yokohama-shi, Kanagawa 240-8501, Japan}

\author[5]{Juan Uriagereka} 
\affil[5]{Department of Linguistics and School of Languages, Literatures and Cultures, 1401 Marie Mount Hall, University of Maryland, College Park, MD 20742-7505, USA}

\maketitle

{  \centerline{\emph{To Aoi, in memory of her loving father}}}

\begin{abstract}

Matrix syntax is a formal model of syntactic relations in language. The purpose of this paper is to explain its mathematical foundations, for an audience with some formal background. We make an axiomatic presentation, motivating each axiom on linguistic and practical grounds. The resulting mathematical structure resembles some aspects of quantum mechanics. Matrix syntax allows us to describe a number of language phenomena that are otherwise very difficult to explain, such as linguistic chains, and is arguably a more economical theory of language than most of the theories proposed in the context of the minimalist program in linguistics. In particular, sentences are naturally modelled as vectors in a Hilbert space with a tensor product structure, built from $2 \times 2$ matrices belonging to some specific group.  Ê

\end{abstract}

\newpage 
\enlargethispage{1cm}
\renewcommand{\baselinestretch}{0.75}\normalsize
{\footnotesize \tableofcontents}
\renewcommand{\baselinestretch}{1.0}\normalsize

\clearpage

\section*{\huge Introduction} 
\addcontentsline{toc}{section}{\protect\numberline{}Introduction}%
\label{sec1}

The currently dominant paradigm in the study of natural language, the so-called minimalist program (MP) developed by Noam Chomsky and his collaborators \cite{mp}, is deeply rooted to earlier models of generative grammar. The aim of this research program  is to better understand language in terms of several fundamental questions. For instance: Why does language have the properties it has? Or what would be an optimal (generative) theory of language?

In accord with its tradition, MP assumes that there is only one human language, though many possible externalizations (which we may think of as Spanish, English, Chinese, etc., as well as sign languages or other modalities). This unique, underlying, human language obeys the conditions of a biologically specified Universal Grammar (UG) and constitutes, somehow, an optimal computational design. UG is considered the 1st factor in language design; socio-historical variation is the 2nd factor; and optimal computational design is referred to as the 3rd factor. Language as a phenomenon is understood to arise and change according to the intricate interactions of these three factors. It is assumed that the faculty of language arose and stabilized within anatomically modern humans, and subsequent changes (from one given language to a historical descendant) are sociological, not evolutionary events that could be biologically transmitted. It is also assumed that epigenetic conditions (input data for language learners) play a central role in the individual development of a given language in an infant's mind; absence of such triggering experience results in ``feral", basically non-linguistic, conditions. 

Against that general background,  it is becoming increasingly clearer that advanced mathematics may help us describe nuanced aspects of language from a theoretical standpoint. A variety of researchers over the last couple of decades have invoked concepts such as groups, operators, and Hilbert spaces, to provide an understanding for the structural conditions observed in language. Interestingly, most of those efforts have sprung not from linguistics, or if they were linguistically-informed, they typically have been connectionist in nature --- and in that, at odds with the generative tradition that has dominated contemporary linguistics \cite{mathphys}. Matrix syntax (MS) can be seen as another such attempt, although it is designed from inception within standard presuppositions from the Computational Theory of Mind (CTM), and more specifically transformational grammar \cite{transgram}. As we understand it, MS is a model of grammar based on a minimal set of axioms that, when combined with linear algebra, go far in accounting for complex long-range correlations of language. 

For example, MS allows us to model linguistic chains (the sets of configurations arising in passive or raising transformations, for instance) by using sums of vectors in some Hilbert space, which is also constructed from the axioms of the model. This is in direct analogy to the superposition of quantum states in quantum mechanics, as we explain below. We have found that the elegant mathematical structure of MS resembles the formal aspects of quantum theory. We are not claiming to have discovered quantum reality in language. The claim, instead, is that our model of language, which we argue works more accurately and less stipulatively than alternatives, especially in providing a reasonable explanation of the behavior of chains, turns out to present mathematical conditions of the sort found in quantum mechanics. In our view, this is quite striking and merits public discussion. 

The model of MS has been introduced in several talks and courses, as well as in an introductory chapter \cite{introms}, but to linguistics audiences. While the linguistic details of the theory will appear in an upcoming monograph \cite{mono}, our purpose here is to present the mathematical details of the system, for an audience with the relevant background. We make an axiomatic presentation of the model, motivating each axiom on the grounds of observed language phenomena, practicality, and optimality. We provide concrete examples, but without entering too much into the (standard) linguistic details, which are being elaborated in the monograph. In this paper we simply let the mathematics speak for itself, as a way to account for certain formal conditions of MP, which we are simply assuming and taking as standard. 

The paper is organized into three parts: Part \ref{p1} (``From words to groups") starts in Sec.\ref{sec2}, where we motivate MS from the perspective of both physics and linguistics. A summary of all the axioms is presented in Sec.\ref{sec3}, with a very brief discussion. Then Sec.\ref{sec4} introduces our ``fundamental assumption" and the ``Chomsky matrices", allowing us to represent lexical categories (noun, verb...) as $2 \times 2$ diagonal matrices. Sec.\ref{sec5} deals with the MERGE operation in linguistics, which we model as matrix product for ``first MERGE", and as tensor product for ``elsewhere MERGE". In Sec.\ref{sec6} we consider self-MERGE at the start of a derivation, and provide our ``anchoring" assumption that only nouns self-MERGE. The different head-complement relations, and linguistic labels, are considered in Sec.\ref{sec7}, analyzing what we call the ``Jarret-graph" and a representation of its dependencies in terms of $2 \times 2$ matrices. Sec.\ref{sec8} studies the group structure of the matrices found, which we call  the ``magniÞcent eight" group or $G_8$, as it is isomorphic to the group ${\mathbb Z}_2 \times {\mathbb Z}_4 $. Those sections are all essentially preliminary, although they already have empirical consequences. 

Using the group $G_8$, in Part \ref{p2} (``From groups to chains") we construct a Hilbert space in Sec.\ref{sec9}, which we call $\mathcal{H}_8$, and which turns out to be isomorphic to the Hilbert space ${\mathbb C}_2$ of a quantum two-level system, i.e., it is a linguistic version of the qubit. The problem of linguistic chains is then addressed in Sec.\ref{sec10}, which we model as sums of vectors in some Hilbert space. Issues on the valid grammatical states and the need for matrix compression in some instances is discussed in Sec.\ref{sec11}. Then, in Sec.\ref{sec12} we introduce reflection symmetry, which allows us to enlarge the structure of the matrices in the model. In particular, we explain the details of what we call the ``Chomsky-Pauli" group $G_{cp}$, which includes the Pauli group $G_p$ as a subgroup, and which allows us to construct a four-dimensional Hilbert space $\mathcal{H}_{cp}$. 

In Part \ref{p3} (``Chain conditions in more detail") we delve into the construction of chains in our formalism. In Sec.\ref{sec12b} we dig a bit deeper into the mathematical structure of the matrices resulting from chain conditions and define the concept of \emph{unit matrices}. Then, we provide a specific example in full detail in Sec.\ref{newsec}. Finally, in the last section we discuss some open questions and extra thoughts, summarize our conclusions, and discuss future perspectives.

\clearpage 

\part{From words to groups}
\label{p1}

\section{Why MS?}Ê
\label{sec2}

MS makes use of many aspects of linear algebra, group theory, and operator theory. Below we show how the resulting model mathematically resembles key aspects of quantum mechanics and works well in the description of many recalcitrant linguistic phenomena, particularly long-range correlations and distributional co-occurrences. In a broad sense, we have two motivations for our model: one for physicists and mathematicians, and one for linguists, although in the end each kind of motivation fuels the other when considered together. 

The simple motivation for MS from the perspective of physics is that its mathematical formalism strikingly resembles some aspects of quantum mechanics. Many familiar mathematical structures used to describe the physics of quantum systems also show up here. For instance: Pauli matrices, $2$-dimensional ($2d$) Hilbert spaces, superpositions, tensor products, and more. More precisely, by assuming a small set of well-motivated axioms, all these mathematical structures follow. It is philosophically very intriguing that this mathematical scaffolding for language --- which is elegant in its own way --- should resemble quantum theory in any sense. From a formal point of view, the theory is particularly appealing in that it involves well-understood objects in mathematics, including several abelian and non-abelian groups. More speculatively, one needs to wonder whether these conditions affect the 3rd factor of language design, and if so what that may tell us about the nature of language as a complex system.

From the perspective of linguistics, MS grew out of a desire to understand the long-range correlations called ``chain occurrences", as arising in sentence \ref{ls1}(a):
\beqa
\label{ls1}
&a.& {\rm Alice ~ seems ~ to ~ like ~ Bob.} \\Ê 
&b.& {\rm Alice ~ likes ~ Bob.} \nonumber
\eeqa
In some sense, ``Alice" in Sen.\ref{ls1}(a) is meant as the ``liker" of Bob, just as it is in Sen.\ref{ls1}(b). At the same time, ``Alice" in Sen.\ref{ls1}(a) is also the subject \emph{of the main clause} (after we introduce ``seems to"). We take ``Alice" to start the derivation in pretty much the same conditions as it does in Sen.\ref{ls1}(b), but then it ``moves" to the matrix subject position, to the left of ``seems". The issue is precisely how this transformation happens and what it entails for the structure and its interpretation.

Linguists call the correlation between the occurrence of \emph{Alice} as the (thematic) subject of \emph{like Bob} and the occurrence of \emph{Alice} as the (grammatical) subject of \emph{seems to like Bob} a ``chain". In a sense, \emph{Alice} is ``in two places at once", although only one of the two occurrences is pronounced. In English, typically the highest occurrence (at the beginning of the sentence) is the only one that can be pronounced, though this is not universally necessary. Thus in Spanish one can say Sen.\ref{ls1}(a) as in Sen.\ref{ls2} (though the order in Sen.\ref{ls1} is fine too):
\beqa
\label{ls2}
&a.& {\rm Parece ~ Alicia ~ apreciar ~ mucho ~ a ~ Roberto. ~ \emph{(Spanish)} } \\Ê
&b.& {\rm Seems ~ Alice ~ appreciate ~ much ~ to ~ Bob. ~ \emph{(Literal  English translation)}} \nonumber \\Ê
&c.& {\rm Alice ~ seems ~ to ~ appreciate ~ Bob ~ a ~ lot. ~ \emph{(Appropriate English translation)}} \nonumber
\eeqa
The fact that \emph{Alice} is realized in one place only (despite existing in two different syntactic configurations as the derivation unfolds) extends to another syntactic interface:  interpretation. To see this, consider a slightly more complex version of Sen.\ref{ls1}, as in Sen.\ref{ls3}:
\beq
{\rm Friends ~ of~ Alice ~seem ~to ~me ~to ~like ~Bob.}
\label{ls3}
\eeq
As is customary in language, with its recursive character, if \emph{Alice} can do something (e.g., move as discussed), a more complex phrase of the same kind containing \emph{Alice} should be able to do the same. In that sense Sen.\ref{ls3}  isn't surprising, nor is it the fact that the movement can take place across a more complex domain (e.g., adding \emph{to me}). 

But now consider some ensuing possibilities:
\beqa 
\label{ls4}
&a.& {\rm Bob ~ and ~ Alice ~ seem ~ to ~ each ~ other ~ to ~ like ~ jazz.} \\Ê
&b.& {\rm Each ~ other's ~ friends ~ seem ~ to ~ Bob ~ and ~ Alice ~ to ~ like ~ jazz.} \nonumber
\eeqa
What is interesting about Sen.\ref{ls4} is that it contains the anaphor \emph{each other}, which requires a plural antecedent. In Sen.\ref{ls4}(a) there is a possible antecedent for \emph{each other}, namely \emph{Bob and Alice}. Matters in Sen.\ref{ls4}(b)  are subtler. It is not so obvious in this instance how \emph{Bob and Alice} can serve as the antecedent of \emph{each other} if we consider these phrases in their ``surface" configurations. However, if we consider \emph{each other('s friends)} in its state prior to movement, then it does seem possible for it to have \emph{Bob and Alice} as its antecdent: 
\beq
\label{ls5}
{\rm Seem ~ to ~ Bob ~ and ~ Alice ~ each ~ other's ~ friends ~ to ~ like ~ jazz.}
\eeq

Finally, the somewhat more complex expressions as in Sen.\ref{ls6} provide the crucial test case for us: 
\beq
\label{ls6}
\small
{\rm Bob ~ and ~ Alice ~ believe ~ a ~ gift ~ of ~ themselves ~ to ~ each ~ other ~ to ~ seem ~ to ~ Tom ~ and ~ Jerry ~ to ~ be ~ acceptable.}
\eeq
It is straightforward to interpret this sentence as having a meaning in which Bob and Alice serves as both the antecedent of \emph{themselves} and \emph{each other}. That is, Sen.\ref{ls6} can reflect the thought that Bob finds giving himself to Alice to be acceptable to Tom and Jerry and Alice finds giving herself to Bob to be acceptable to Tom and Jerry. Alternatively, Tom and Jerry can serve as the antecedent of both anaphors, in which case the intended interpretation is that Bob and Alice believe it to seem to Tom that giving himself to Jerry is acceptable and they also believe it to seem to Jerry that giving himself to Tom is acceptable. But there is also an unavailable reading: it is impossible to interpret Sen.\ref{ls6} as meaning that Bob and Alice believe that a gift of themselves (= Bob and Alice) to Tom to seem acceptable to Jerry and that a gift of themselves to Jerry to seem acceptable to Tom. In other words, while the anaphors \emph{themselves} or \emph{each other} can take either the matrix subject \emph{Bob and Alice} or the experiencer \emph{Tom and Jerry} as an antecedent --- not surprising given the discussion in the previous paragraph --- it seems impossible to get one of the anaphors to take as an antecedent \emph{Bob and Alice} and the other anaphor to take as antecedent \emph{Tom and Jerry}. In order to prevent such a reading, the phrase \emph{a gift of themselves to each other} should not get an interpretation in both configurations that it occupies as the derivation unfolds. 

A chain can be thought of as the set of configurations in which we find multiple occurrences of a moved element in a sentence. Just as only one of these occurrences can be pronounced (we cannot say, grammatically, *\emph{Alice seems Alice to like Bob}), only one of the occurrences in a chain can be interpreted. This of course can be stipulated --- but it makes no formal sense. We could have equally stipulated, say, that all chain occurrences must be interpreted, or only half, or any other bizarre statement, one stipulation being as good as any other. One of our main goals in this paper is to develop a formal system in which it follows as a necessity that only one among the chain occurrences gets pronounced and interpreted. We know of no current worked out explanation of that fact\footnote{ For example, Collins and Stabler \cite{Collins} define the notion ``occurrence", stipulating a reduction of multiple occurrences to only one as seems necessary on empirical grounds. That stipulation, however, is what needs to be explained, since the very same system could have introduced any other stipulation, or not involve occurrences at all. \color{black}}. 

Coming back to Sen.\ref{ls1}: there are two occurrences of \emph{Alice} in Sen.\ref{ls1} --- it is, at some point in the derivation, simultaneously in two different places. Importantly, only one of those occurrences of \emph{Alice} is ``externalized" when the sentence is pronounced and, as we have argued, only one is interpreted (``internalized" as meaning) as well, although these do not necessarily have to be the same occurrences. 

We will argue that chains are best described as sums of vectors in some Hilbert space with a tensor product structure and thus act very much like superposition does in quantum mechanics. In this sense, chains are the analogue of ``Schr\"odinger's cat" in quantum physics. As the syntactic derivation is sent to an interface (for phonetic externalization or meaning in the shape of a logical form), one of the configurational chain-occurrence options is randomly chosen, as far as we know, which is very much analogous to the quantum-mechanical ``collapse" of a quantum state after a measurement. At that point externalization/interpretation could be seen as the literal collapse of a syntactic state, in terms familiar to physicists. It turns out that situations of this sort, and even more complex ones, are ubiquitous in language: passives, interrogatives, relative clauses, and much more. To date there is no satisfactory explanation of why things are the way they seem to be, among many other imaginable alternatives. This is where MS naturally encompasses a vector space able to describe the odd behavior of chain occurrences. 

\section{Teaser of the axioms} 
\label{sec3}

We will base MS on a dozen axioms. We will present all of them together, so as to have an overall idea of what is coming. In subsequent sections we elaborate on the motivations for each, as well as various consequences. The axioms are as follows:

\bigskip
$\bullet$ \emph{{\bf Axiom 1 (fundamental assumption):}Ê lexical attributes are $N=1$ and $V=i$.}Ê

\bigskip
$\bullet$ \emph{{\bf Axiom 2 (Chomsky matrices):}Ê the four lexical categories are equivalent to the diagonal matrices}Ê
\beq
{\rm Noun} = 
\begin{pmatrix}Ê
1 & 0Ê\\Ê
0 & -iÊ
\end{pmatrix}Ê
~~
{\rm Verb} = 
\begin{pmatrix}Ê
- 1 & 0Ê\\Ê
0 & iÊ
\end{pmatrix}Ê
~~~
{\rm Adjective} = 
\begin{pmatrix}Ê
1 & 0 Ê\\Ê
0 & iÊ
\end{pmatrix}Ê
~~~
{\rm Elsewhere} = 
\begin{pmatrix}Ê
- 1 & 0Ê\\Ê
0 & -iÊ
\end{pmatrix}. 
\label{ax2}
\eeq

$\bullet$ \emph{{\bf Axiom 3 (multiplication):}Ê 1st MERGE (M) is matrix multiplication.}Ê

\bigskip
$\bullet$ \emph{{\bf Axiom 4 (tensor product):}Ê elsewhere M is matrix tensor product.}Ê

\bigskip
$\bullet$ \emph{{\bf Axiom 5 (Kayne):} M is anti-symmetrical.}Ê

\bigskip
$\bullet$ \emph{{\bf Axiom 6 (Guimar$\tilde{{\rm {\bf a}}}$es):} only nouns self-M.}Ê

\bigskip
$\bullet$ \emph{{\bf Axiom 7 (determinant/label):}Ê the linguistic label of a phrase is the complex phase of the matrix determinant.}Ê

\bigskip
$\bullet$ \emph{{\bf Axiom 8 (Hilbert space):}Ê linguistic chains are normalized sums of vectors from a Hilbert space.}Ê

\bigskip
$\bullet$ \emph{{\bf Axiom 9 (interface):}Ê when chain $\ket{\psi}$ is sent to an interface, its vector gets projected in one of the elements $\ket{\phi}$ being specified according to the probability $|\braket{\phi}{\psi}|^2$.}Ê       

\bigskip
$\bullet$ \emph{{\bf Axiom 10 (filtering):}Ê valid chains are separable (non-entangled) states in the Hilbert space.}Ê  

\bigskip
$\bullet$ \emph{{\bf Axiom 11 (compression):}Ê matrices of valid grammatical chains have a non-trivial null (irrelevant) eigenspace.}Ê  Ê

\bigskip
$\bullet$ \emph{{\bf Axiom 12 (structure):} Êthe system has reflection symmetry.}

As we discuss below, Axiom 12 is what allows us to produce a structure in terms of what we call the ``Chomsky-Pauli group", which in turn includes the ``Pauli group". One may elaborate the full theory of MS without this axiom. However, in the rough sense in which the universal phonetic alphabet is effectively a periodic table of linguistic sounds, in some practical situations more structure may be required, and this is why we include it at the end. It is in fact the simplest way to obtain a very large and reasonable ``periodic table" of grammatical elements out of the model. Some of the axioms may follow from deeper conditions and relate in other ways, but this is not the focus of the current paper. As noted, we know of no alternative to the present system that can explain the odd behavior of chains described above. We devote the rest of the paper to showing, step by step, how these axioms work, why they are useful in the study of language, and what their mathematical similitudes are with known physical concepts. 

\section{Fundamental assumption and Chomsky matrices}
\label{sec4}

ÊFollowing ideas going back to P$\ddot{{\rm a}}$nini, \color{black} Noam Chomsky in 1955 \cite{chomsky1955} proposed categorizing lexical items into four types: noun (N), verb (V), adjective (A) and elsewhere (P). The last one of these categories includes anything not included in the first three, i.e., essentially prepositions, postpositions (i.e., adpositions), and other such categories. In 1974 \cite{chomsky74}, he further proposed a way to state commonalities among the four major categories, in terms of abstract features. This followed the intuition from classical phonological theory of describing phonemes as lists of attributes \cite{phono}, for instance: 
\beq
[ {\rm d } ] = 
\begin{pmatrix}Ê
+ {\rm consonantal}Ê\\Ê
+ {\rm voiced}Ê\\Ê
+ {\rm alveolar}Ê\\Ê
+ {\rm stop}
\end{pmatrix}.Ê
\eeq 
In the left side of the equation we have the phoneme corresponding to the sound typically written with the letter d in English, and in the right side a list of its phonological attributes. As such, this is actually a matrix, or more precisely, a vector or $1d$ array of attributes. Chomsky extended this general idea to lexical categories too. In particular, he proposed the following correspondence between the four basic lexical categories and lists of attributes: 
\beq
{\rm Noun} = 
\begin{pmatrix}Ê
+ NÊ\\Ê
- VÊ
\end{pmatrix}Ê
~~
{\rm Verb} = 
\begin{pmatrix}Ê
- NÊ\\Ê
+ VÊ
\end{pmatrix}Ê
~~~
{\rm Adjective} = 
\begin{pmatrix}Ê
+ NÊ\\Ê
+ VÊ
\end{pmatrix}Ê
~~~
{\rm Elsewhere} = 
\begin{pmatrix}Ê
- NÊ\\Ê
- VÊ
\end{pmatrix}.
\label{ch}
\eeq
In Eq.(\ref{ch}), $\pm N$ and $\pm V$ are meant as abstract lexical attributes corresponding to a variety of syntactic phenomena. For instance, $-N$ categories (verbs and prepositions) assign case, whereas $+N$  categories (nouns and adjectives) receive case instead. Also, $+V$ categories (verbs and adjectives) typically denote properties or actions, whereas $-V$ categories typically denote entities or locations. It is important not to confuse the lexical attribute $V$ with the lexical category $V$ (which has the lexical attribute $V$ in the positive and the lexical attribute $N$ in the negative). 

Taking Chomsky's proposal very seriously, we express it via diagonal matrices with numerical attributes. This is done by means of the first two axioms: 

\bigskip
$\bullet$ \emph{{\bf Axiom 1 (fundamental assumption):}Ê lexical attributes are $N=1$ and $V=i$.}Ê

\bigskip
$\bullet$ \emph{{\bf Axiom 2 (Chomsky matrices):}Ê the four lexical categories are equivalent to the diagonal matrices}Ê
\beq
{\rm Noun} = 
\begin{pmatrix}Ê
1 & 0Ê\\Ê
0 & -iÊ
\end{pmatrix}Ê
~~
{\rm Verb} = 
\begin{pmatrix}Ê
- 1 & 0Ê\\Ê
0 & iÊ
\end{pmatrix}Ê
~~~
{\rm Adjective} = 
\begin{pmatrix}Ê
1 & 0 Ê\\Ê
0 & iÊ
\end{pmatrix}Ê
~~~
{\rm Elsewhere} = 
\begin{pmatrix}Ê
- 1 & 0Ê\\Ê
0 & -iÊ
\end{pmatrix}. 
\label{ax2}
\eeq
We call these the Chomsky matrices. The reason why we choose to codify Chomsky's four basic lexical elements in Eq.(\ref{ax2}) as these diagonal matrices and not, e.g., as column vectors, is because we find it more natural to operate with matrices than with vectors, as we see in the forthcoming sections. Moreover, it will be convenient to introduce structures which are represented by antidiagonal matrices, as we will see when discussing Axiom 12. 

Axiom 1 stems from a deep intuition behind Chomsky's 1974 decision to represent all four categories in terms of two ``conceptually orthogonal" features. These ``binary oppositions" go back to the notion ``distinctive feature" developed in Ref.\cite{jak}. Distinctive features as such are binary (+ or -), but in addition the features can be classified among themselves; for example, among the consonantal phonemes, these could be stop or continuant, while among the vocalic phonemes, they could be tense or lax, etc. Intuitively, the difference between phonemes being vocalic or consonantal is far greater than the differences occurring within those dimensions. So too with lexical categories: $N$ and $V$ are lexical attributes that cannot be more different, and hence it makes mathematical sense to codify them as $1$ and $i$ respectively (or in any other orthogonal fashion). Although this is purely an assumption, we argue that it is justified by the fact that it allows us to account for well-known distributional properties that we return to. 

\section{MERGE: multiplication and tensor product}
\label{sec5}

\begin{figure}
	\centering
	\includegraphics[width=0.18\linewidth]{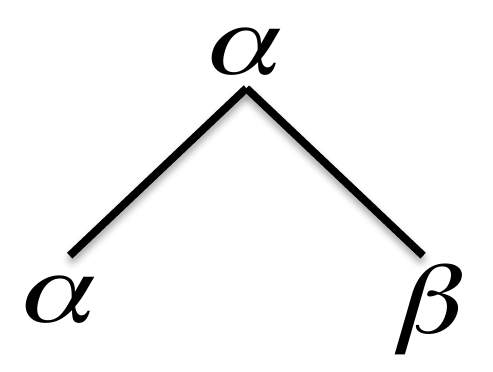}
	\caption{ÊMERGE operation in linguistics: two lexical elements $\alpha$ and $\beta$ producing a new element with the properties of $\alpha$. For instance, the verb ``eat" merges with the noun ``pasta" to form the verb phrase ``eat pasta". As such, this is also an example of first-MERGE, since there is a head-complement relation between the bare lexical element ``eat", which is the head, and ``pasta", which acts as its complement. We refer also to the physical interpretation of MERGE in Ref.\cite{gallegoorus}, which will come up later in Sec.\ref{sec12b}. \color{black}}Ê
	\label{fig2}
\end{figure}

We now consider the M operation in linguistics, proposed by Chomsky as the basic pillar to build up linguistic structures \cite{merge}. In M, two syntactic objects are combined to form a new syntactic unit. The operation can be applied recursively, thus having the ability to create different generations of units. For instance, a verb (like \emph{eat}) and a noun phrase (like \emph{pasta}) may undergo M to produce a combined object described as a verb phrase (\emph{eat pasta}). The diagrammatic explanation is shown in Fig.\ref{fig2}. Despite M being the basic structure building operation, there are, in fact, different types of M. The distinction between external M and internal M (the latter also being called ``movement") is commonly mentioned in the MP literature, and will be reviewed in Sec.\ref{sec10} when talking about chain formation. The distinction between 1st M as opposed to elsewhere M is subtler, and sometimes only made implicitly. Axiomatically, for us, the distinction is as follows: 

\bigskip
$\bullet$ \emph{{\bf Axiom 3 (multiplication):}Ê 1st MERGE (M) is matrix multiplication.}Ê

\bigskip
$\bullet$ \emph{{\bf Axiom 4 (tensor product):}Ê elsewhere M is matrix tensor product.}Ê
\bigskip

\begin{figure}
	\centering
	\includegraphics[width=0.8\linewidth]{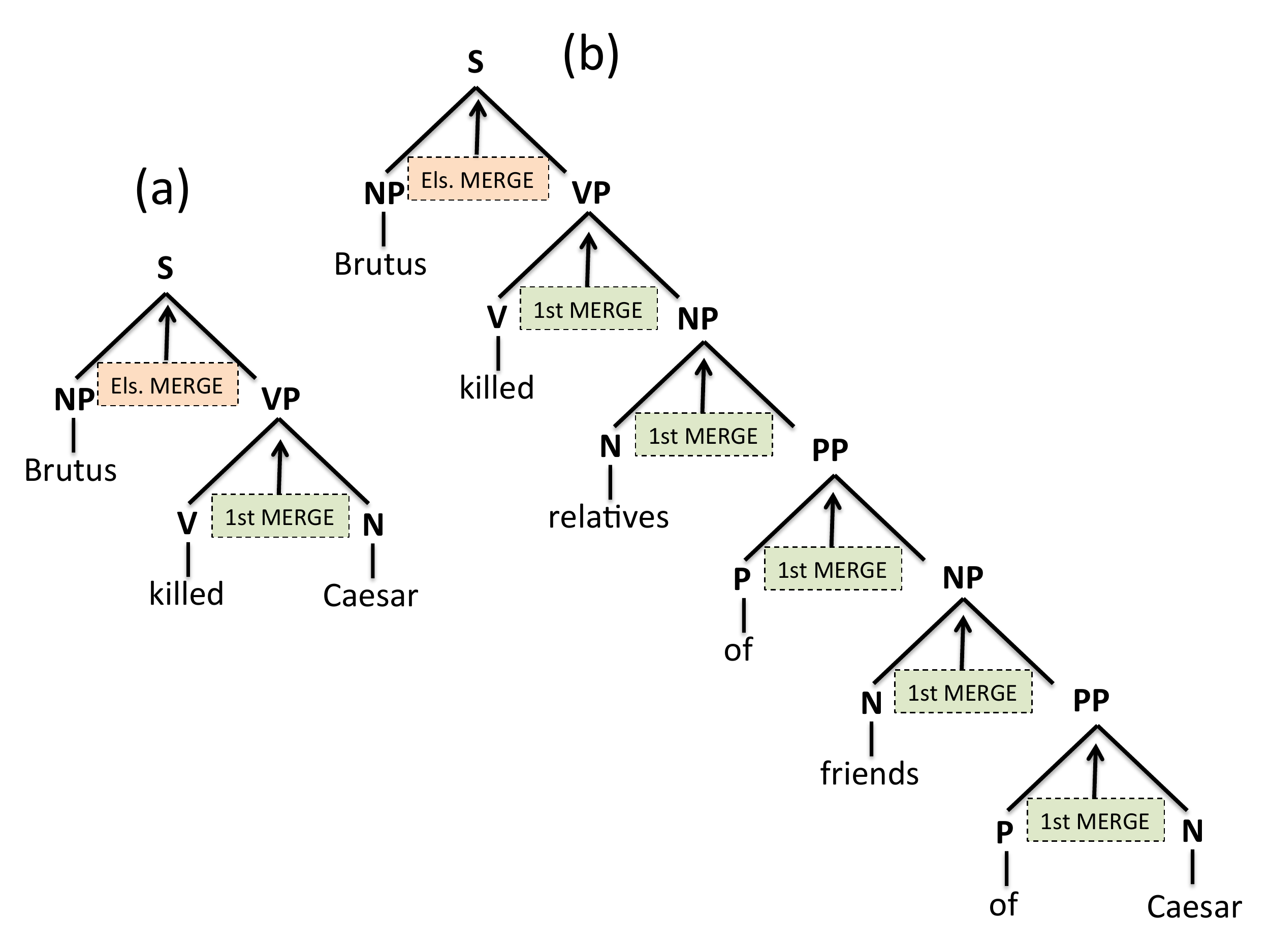}
	\caption{Simplified syntactic trees, in the usual linguistic style, of (a) ``Brutus killed Caesar", and (b) ``Brutus killed relatives of fiends of Caesar". First and elsewhere M operations are highlighted. For expository purposes we avoid ``grammatical categories" like Tense, Number, and so on, and we do not dwell, yet, on the situation arising at the very bottom of the tree, as in the next section.}
	\label{fig3}
\end{figure}

1st M denotes the merger of a bare lexical item from the mental lexicon (e.g., a verb) --- which plays the nuclear role of a ``head" that is said to ``govern" what it merges with --- together with its ``complement", typically a complex object that has already been assembled in the syntactic derivation. Note the asymmetry here: the atomic head comes from the lexicon, while the complement that this head governs is already in the derivation, and it may be arbitrarily complex. In linguistic jargon, this operation results in what is called a ``projection". Take for instance the sentence \emph{Brutus killed Caesar}. Here, the verb \emph{killed} and the noun phrase \emph{Caesar} are merged via 1st M into the verb phrase \emph{killed Caesar}, which turns out to be the predicate of the whole sentence, see Fig.\ref{fig3}(a). Playing with this idea one can in principle build long recursive relations that could describe Brutus's actions, as in \emph{... killed the Emperor of Rome, ... killed relatives of the Emperor of Rome, ... killed rumors about relatives of the Emperor of Rome}, etc.; see Fig.\ref{fig3}(b). In all these instances we say that the head verb \emph{killed} governs its complement noun phrase ($NP$) and that the consequence of this merger is the projection of the verb to yield what is called a verb phrase ($VP$). 

Complementary to this, elsewhere M denotes the merger of two syntactic objects both of which are the results of previous mergers in the derivation (in linguistic jargon, one is merging ``two projections"), for instance, the merger of an $NP$ (like \emph{Brutus}, or \emph{Caesar's son}, or \emph{Caesar's son to Servilia}) and a $VP$ yielding the semantic ingredients for a whole sentence S. Just as the phrase undergoing 1st M to a head is called its complement, a phrase elsewhere merged to the projection of a head-complement relation is called a specifier. 

As we show below, 1st and elsewhere M result in structures with rather different linguistic properties. Here are a couple of clear ones for perspective:
\beqa
\label{s11}
 &a.& {\rm Sharks ~often ~go ~killing ~seals.} \\Ê
 &b.& {\rm Sharks ~often ~go ~\text{seal-killing}.} \nonumber \\Ê
 &c.& {\rm *Often ~go ~\text{shark-killing} ~seals.} \nonumber 
\eeqa
Complements can incorporate to a head verb as in Sen.\ref{s11}(b) (with the basic underlying structure in Sen.\ref{s11}(a)), but specifiers cannot incorporate to a head verb as in Sen.\ref{s11}(c) --- which makes as much semantic sense as Sen.\ref{s11}(a) but is ungrammatical. Similarly, sub-extraction of an information question from a complement is possible as in Sen.\ref{s12}(b) (which can be answered as in Sen.\ref{s12}(a)), but a similar sub-extraction involving a specifier is impossible as in Sen.\ref{s12}(c) (which ought to be able to also obtain Sen.\ref{s12}(a) as an answer, but is ungrammatical):
\beqa
\label{s12}
&a.& {\rm Friends ~of ~the ~mafia ~occasionally ~kill ~enemies ~of ~politicians.} \\Ê
&b.& {\rm Who ~do ~friends ~of ~the ~mafia ~occasionally ~kill ~enemies ~of ?} \nonumber \\Ê
&c.& {\rm *Who ~do ~friends ~of ~occasionally ~kill~ enemies ~of ~politicians?} \nonumber
\eeqa

The differences reported above are syntactic, but one can also observe interesting semantic asymmetries between complements and specifiers:
\beq
{\rm In ~Delos, ~humans ~decimated ~goats.} 
\label{s13}
\eeq
Delos is an island where humans managed to decimate the main fauna, which eventually came to be uninhabited. Importantly for the matter at stake, the presence of both goats (complement in this context) and humans (specifier) in Delos was transient. However, in a sentence like Sen.\ref{s13}, what measures the duration of the \emph{decimating} event is the existence of the goats denoted by the complement, not the existence of the humans denoted by the specifier. It could have been that humans actually had ceased to exist in Delos before the goats had been decimated, whereas for the decimation event to be completed, it is crucial that all goats in Delos disappear. 
 
Given these and other such distinctions between the two types of M, our Axioms 3 and 4 separate each by way of a different mathematical operation. The reasons why these two forms of (external) M are handled in this way will become clear in the following sections. Also, as explained in Ref.\cite{gallegoorus}, M can be understood as the linguistic equivalent of a coarse-graining process in physics, where what is being coarse-grained is linguistic information, the coarse-graining entailing different time scales\footnote{ÊA coarse-graining is, by definition, the process that keeps the relevant degrees of freedom to describe a system at a given scale, while throwing away non-relevant ones. As an example, think of it as ``zooming out" the pixel-by-pixel description of an image (lots of information codified in individual bits), in order to see the overall image (e.g., ``a tree").  In physics, the mathematical framework encompassing these ideas is called ``Renormalization". As explained in Ref.\cite{gallegoorus}, language itself has a renormalization structure, where complex structures (e.g., sentences) \emph{emerge} at long time scales from fundamental units (e.g., words) and the correlations amongst them. \color{black}}. In 1st M, matrix multiplication involves loss of linguistic information: one cannot uniquely determine the original matrices being multiplied if one is given their product only. For elsewhere MERGE, the situation is subtler, and the coarse-graining picture emerges from the number of relevant eigenspaces in the resulting matrix, as will be discussed in Sec.\ref{sec11}.

\section{Anti-symmetry and the syntactic anchor}Ê
\label{sec6}

Lest this be confusing in the examples above, note that plural nouns or names, as such, are commonly understood as having internal structure and are therefore treated not as head nouns, but as noun phrases ($NP$) themselves. While this may not be obvious in English (where we say \emph{Brutus}), it is more so in languages such as German or Catalan (where we could say \emph{der Brutus} and \emph{el Brutus}). Under this common assumption, M between \emph{Brutus} and \emph{killed Caesar}  in Fig.\ref{fig3}(a) is actually an instance of elsewhere (not 1st) M. In this sense we may find a confusing situation as in Sen.\ref{s14}, which readers need to be alerted to:
\beqa
\label{s14}
&a.& {\rm Caesar~ offended ~several ~{\bf relatives} ~of ~Brutus.} \\
&b.& {\rm {\bf Relatives} ~can ~be ~a ~nuisance.} \nonumber \\
&c.& {\rm Oliver ~Twist ~lacked ~ {\bf relatives}.} \nonumber
\eeqa
The boldfaced {\bf relatives} is quite different in these circumstances. The merger between ``relatives" and ``of Brutus" in Sen.\ref{s14}(a) is a 1st M because of the head-complement relation: ``of Brutus" is the complement of ``relatives" there. In contrast, the merger between the specifier ``relatives" and the projected phrase ``can be a nuisance" in Sen.\ref{s14}(b) is an elsewhere M --- which does pose the question of how ``relatives" gets to project so as to be a specifier to begin with. Arguably, this matter relates to the situation in Sen.\ref{s14}(c), where this time around we want ``relatives" to be the complement (of a verb). We return shortly to the latter situation, but the distinctions in Sen.\ref{s14} should suffice to illustrate the interesting behavior of nouns.

Most linguists agree that nouns are special to the linguistic system, perhaps for deep cognitive reasons. It is, for instance, well known that children typically start by acquiring nouns in the so-called ``one word stage" \cite{oneword}. Also, \emph{anomias} \cite{anomia} selectively targeting nouns and specific forms of language break-down targeting the noun system are known --- in ways that do not appear to extend specifically to the verbal, adjectival or prepositional systems. Regardless of what the reasons are for the central role nouns play within grammar, we want the formal system to be sensitive to this fact, which relates to another important assumption, expressed in Axiom 5: 

\bigskip
$\bullet$ \emph{{\bf Axiom 5 (Kayne):} M is anti-symmetrical.}Ê
\bigskip

Since Richard Kayne's seminal work on this matter \cite{kayne}, it is generally assumed that syntax, and more concretely M, is anti-symmetrical. In the case of 1st M the anti-symmetry is evident: one merges a lexical element acting as a head, say a verb $V$, with a phrase constructed from previous mergers, such as an $NP$ acting as a complement. These elements stand in an asymmetric relation: one is an ``atomic" item from the lexicon and the other is not. But this in turn implies that at the very beginning of the derivation, when there are no complex projections/phrases (yet), either the asymmetry breaks down or something must be able to self-merge, so as to construct a phrase. That is the anti-symmetry situation, in that (in linguistic jargon) a relation is generally considered anti-symmetrical if it is asymmetrical except when holding with itself\footnote{  Anti-symmetry (the property of a relation holding asymmetrically except when it holds with itself) holds of relations, and here we are talking about an operation. Still, there are reasons why the merge operation is asymmetrical except when it involves a category merging with itself. Kayne argued that, among words, asymmetric c-command dependencies (see fn. 19) map to precedence relations, thereby linearizing phrase-markers that, otherwise, the system does not organize sequentially. He also noted how this poses a problem for words that merge to one another, as each c-commands the other (symmetrically): such terminal items could not linearize with regards to one another. Max Guimar‹es, in work we return to immediately, noted an interesting sub-case where this does not matter: when the words that merge are identical. That is, if a word X merges to another X to form X-X, it is meaningless to ask which of those X's comes first. The system cannot tell, and it does not matter either. This is the particular sense in which we take Merge to be anti-symmetrical. In general, Merge is asymmetrical so as to satisfy Kayne's linearization requirement; with one exception: when a category merges to itself, resulting in mutual (symmetrical) c-command of the merged items. We assume that in such instances the phonological output of those merged objects is either randomly linearized (as in \emph{No, no...}), or perhaps it is also possible to linearize it in a ``smeared" way, as in \emph{Nnnn...no}, which we take to be equivalent to \emph{No, no,} syntactically and semantically, if not phonologically (similarly, \emph{rats, rats!} would be equivalent to \emph{rrrrrats!}, etc.) \color{black}}. In other words, the initial self-merge starts the derivation. 

Now note: because per Axiom 3, 1st M is matrix multiplication of the Chomsky matrices (Axiom 2), and the initial self-M of any lexical element yields the very same matrix:
\beq
({\rm Noun})^2  = 
\begin{pmatrix}Ê
1 & 0Ê\\Ê
0 & -1Ê
\end{pmatrix}Ê
~
({\rm Verb})^2  = 
\begin{pmatrix}Ê
1 & 0Ê\\Ê
0 & -1Ê
\end{pmatrix}Ê
~
({\rm Adjective})^2  = 
\begin{pmatrix}Ê
1 & 0Ê\\Ê
0 & -1Ê
\end{pmatrix}Ê
~
({\rm Elsewhere})^2  = 
\begin{pmatrix}Ê
1 & 0Ê\\Ê
0 & -1Ê
\end{pmatrix}.Ê
\label{anchoring}
\eeq
The matrix resulting in the same four situations above is the Pauli matrix $Z$. Now, for a semiotic system carrying information between speakers, it is reasonable to expect that \emph{only one of these four options is grammatically possible.} 

Imagine if Paul Revere's famous lantern signal system were coding ``one if by land, two if by land" or ``one if by sea, two if by sea". Obviously, that would not be as informative as Revere's actual message: ``one if by land, two if by sea". We may think of that as a semiotic assumption that we take to be so elementary and general as not to require its explicit statement as an axiom of our particular system. Now whereas the need for such an assumption is obvious, how we implement it is a different issue, so long as some information anchoring exists. 

The anchoring we will assume here is based on a version of an insight by Max Guimar$\tilde{{\rm a}}$es \cite{cuimkane} for the initial step of syntactic derivations, which Kayne later popularized in Ref.\cite{kayne2}:

\bigskip
$\bullet$ \emph{{\bf Axiom 6 (Guimar$\tilde{{\rm {\bf a}}}$es):} only nouns self-M.}Ê
\bigskip

This axiom implies that Pauli matrix $Z$ is, in MS, a noun phrase $NP$. Notice that other mathematical options would have also been logically possible: one could have chosen a system where, say, only verbs are allowed to self-M, or only adjectives, or only prepositions. All these schemes are also valid language models. But in our case, the assumption that seems to match what is observed in human language (i.e., produces a system consistent with observations), is  the self-M of nouns, thereby instantiating a ``privileged" role for these elements within the system.

\section{The Jarret-graph, labels, and determinants}Ê
\label{sec7}

\begin{figure}
	\centering
	\includegraphics[width=0.55\linewidth]{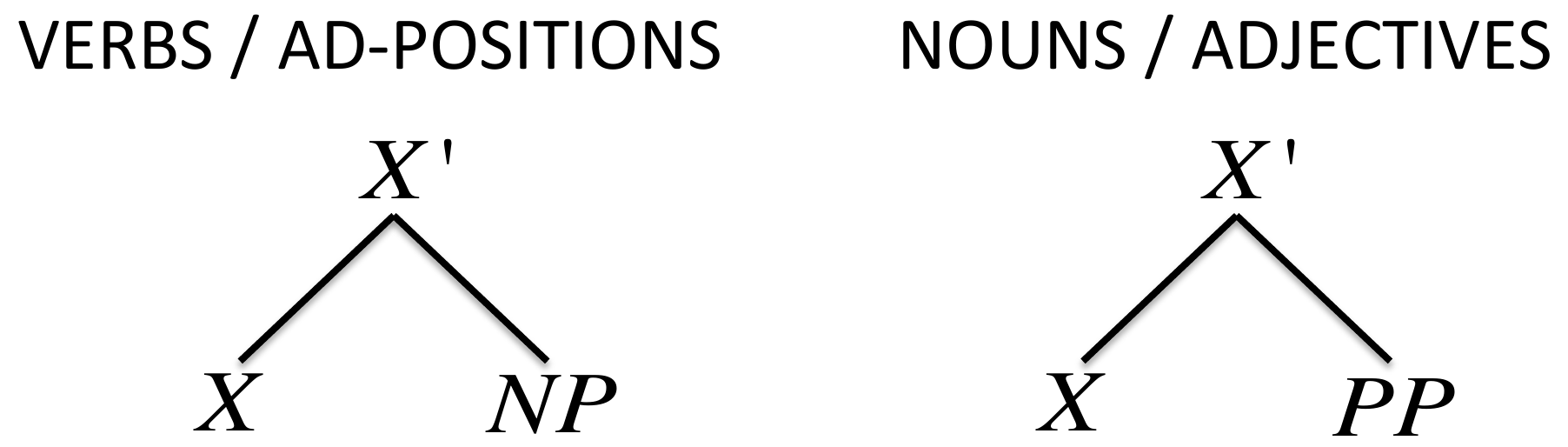}
	\caption{Grammatical selections for the various Chomsky heads.}
	\label{fig34}
\end{figure}

\begin{figure}
	\centering
	\includegraphics[width=0.45\linewidth]{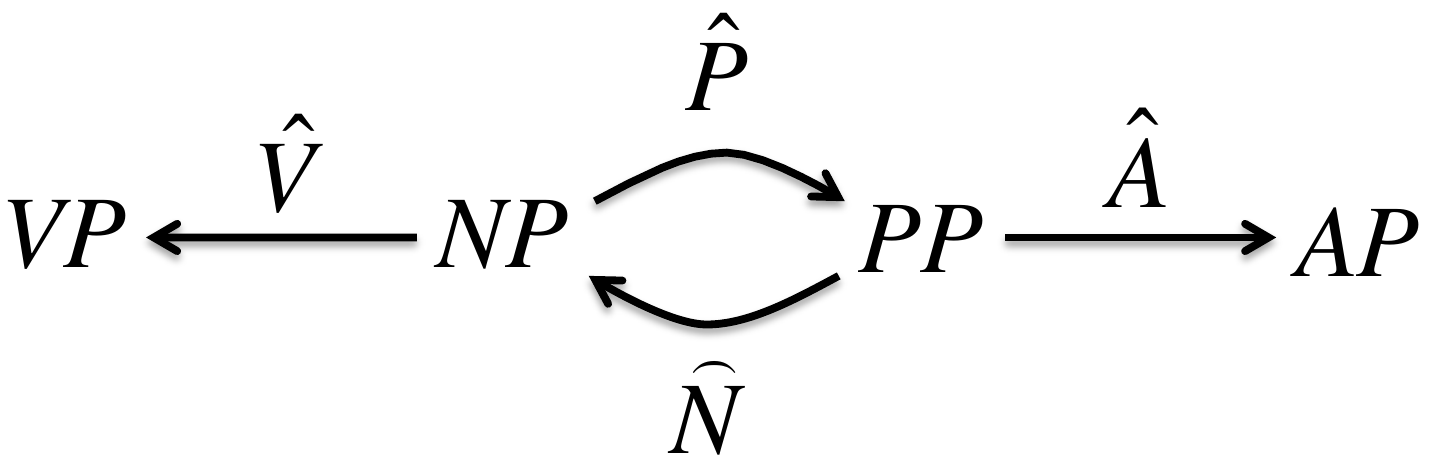}
	\caption{Jarret-graph of head-complement relations via first MERGE. It is a directed graph, where the nodes correspond to phrases (complements), and the links to bare lexical categories (heads). The interpretation is that the heads act as operators on the phrases, producing a new phrase. For instance, $\hat{V}(NP) = VP$, with the ``hat" meaning that it acts as an operator.}
	\label{fig4}
\end{figure}

The next step in our presentation stems from the relatively obvious fact that not all imaginable head-complement relations are grammatical. It is not difficult to summarize the main combinations that do and do not occur, as in Table \ref{tabHC}.
\begin{table}[h]
\centering
\begin{tabular}{  || l || l ||}
\hline 
& \\ 
i. NOUNS & ii. VERBS \\
  & \\
  $a.$ *[portrait Rome]     &                  $a.$   [destroy Rome]  \\    
                     $b.$  *[portrait eat]           &                                    $b.$ *[destroy see]  \\
                    	       $c.$  *[portrait red]    &                                         $c.$  *[destroy happy]          \\
                      $d.$   [portrait of Rome]      &                                $d.$  *[destroy in Rome]  \\
                     & \\\hline
                     & \\
                     iii. ADJECTIVES & iv. PREPOSITIONS \\Ê   
   & \\
   $a.$  *[proud Rome]  &	       $a.$   [in Rome]      \\             
               	                  $b.$ *[proud eat]          &        	                            $b.$ *[in wait]   \\
                              	   $c.$  *[proud angry]        &                   	             $c.$ *[in tall] \\
                              	   $d.$   [proud of Rome]         &                                      $d.$ *[in near (Rome)] \\
   & \\\hline
\end{tabular}
\caption{Different possible head-complement relations.}
\label{tabHC}
\end{table}
Facts (iii) and (iv) in Table \ref{tabHC} are close to 100$\%$ true across languages (ignoring word order, like whether an ``adposition" is pre or post-positional)\footnote{ÊLinguists often like to discuss exceptions to general patterns like the present one. For example, there is such a thing as ``preposition stacking" in languages like English (e.g. \emph{the cafe is up over behind that hill}). It is pretty clear that such situations are marked, however. At this point we are trying to focus on the main patterns found across languages, which are quite overwhelmingly as in Table \ref{tabHC}. \color{black}}. Facts (i) and (ii) are statistically overwhelming: if a noun has a complement, it is virtually always a pre/post-positional phrase ($PP$). Finally, most studied languages have at least 50$\%$ transitive verbs (with an $NP$ complement); half of these languages have 60$\%$ such verbs; in several the proportion goes up to 70$\%$. That means that all other verb classes (including ``intransitive", ``ditranstive", etc.) are ``the rest" of verbal dependencies.  Idealizing the situation in Table \ref{tabHC},  we summarize valid combinations as in Fig.\ref{fig34}. Physicist Michael Jarret \cite{mj} has conceptualized this situation in terms of a graph that, in his honor, we affectionately call the Jarret-graph. The graph is presented in Fig.\ref{fig4}. 

Nodes in the graph represent phrases acting as complements and directed links (arrows) represent bare lexical categories acting as heads. The head, acting as an operator, takes a phrase as an input and produces a new phrase as an output. So the head operators map different types of phrases amongst themselves. 1st M is therefore understood as \emph{the action of head operators on phrases}. In particular, the dependencies coming out from the Jarret-graph are the following:
\beq
\hat{P}(NP) = PP,  ~~~~ \hat{V}(NP) = VP, ~~~~ \hat{N}(PP) = NP, ~~~ \hat{A}(PP) = AP. 
\eeq
In the above equations, $\hat{P}, \hat{V}, \hat{N}$ and $\hat{A}$ are operators for the lexical categories of preposition (or elsewhere), verb, noun and adjective, acting as the ``head" in 1st M. Also, obviously, $NP, PP, AP$ and $VP$ correspond to noun phrase, prepositional phrase, adjectival phrase and verb phrase, respectively. The Jarret-graph is simply a diagrammatic representation of these four equations in terms of a directed graph. 

It may be noticed that, although it is a convenient representation, we have not included the Jarret-graph as one of our axioms. This is because its character is actually not axiomatic, the combinations it presents follow from the interactions of other conditions that we need to continue presenting, together with our anchoring assumptions (Axioms 5 and 6). From those assumptions alone, we may recall, we obtain the Pauli $Z$ as our initially projected (self-merged) $NP$:
\beq
NP = 
\begin{pmatrix}Ê
1 & 0Ê\\Ê
0 & -1Ê
\end{pmatrix}. 
\eeq
Just by plugging in this matrix at the $NP$ node of the Jarret-graph, and also using the matrices in Eq.(\ref{ax2}) to represent Chomsky's head lexical categories acting as operators, we can perform all relevant 1st MÕs understood as matrix products (per Axiom 3). Such products are fairly simple (basically, entry-wise), since all of the matrices are diagonal. As a result, we find something significant: \emph{the outcome of the graph is closed and unique}, as can be seen in Fig.\ref{fig5}.

The Jarret-graph, in combination with the previous axioms, fully fixes a valid representation for all nodes and vertices in terms of eight $2 \times 2$ matrices. The head matrices (operators, links of the graph) are the ones in Eq.(\ref{ax2}), whereas for the complement matrices (nodes of the graph) we always have two options, namely: 
\beqa
 NP = 
\left\{ \begin{pmatrix}Ê
1 & 0Ê\\Ê
0 & -1
\end{pmatrix},  
\begin{pmatrix}Ê
-1 & 0Ê\\Ê
0 & 1
\end{pmatrix} \right\}ÊÊ
&~~~&
 VP = 
\left\{ \begin{pmatrix}Ê
1 & 0Ê\\Ê
0 & iÊ
\end{pmatrix}, Ê
\begin{pmatrix}Ê
-1 & 0Ê\\Ê
0 & -iÊ
\end{pmatrix}\right\}ÊÊ\nonumber \\ 
 PP = 
\left\{ \begin{pmatrix}Ê
1 & 0Ê\\Ê
0 & -i
\end{pmatrix}, 
\begin{pmatrix}Ê
-1 & 0Ê\\Ê
0 & i
\end{pmatrix} \right\}Ê
&~~~&
 AP = 
\left\{\begin{pmatrix}Ê
1 & 0Ê\\Ê
0 & 1Ê
\end{pmatrix}, 
\begin{pmatrix}Ê
-1 & 0Ê\\Ê
0 & -1Ê
\end{pmatrix} \right\}Ê
\label{ax5}
\eeqa
It is not difficult to see what these ``twin" matrices for each lexical category have in common: the distinctive feature is their matrix determinant\footnote{ÊAdditionally, they are Êeach other's additive inverses. \color{black}}. In particular, we see that the determinant for the matrices of $NP$ is $-1$, for $VP$ it is $i$, for $PP$ it is $-i$, and for $AP$ it is $1$. Therefore, we may take the matrix determinant to be the distinctive ``label" that allows us to recognize, numerically, which type of phrase we are operating with. We elevate this to the category of our seventh axiom:

\bigskip
$\bullet$ \emph{{\bf Axiom 7 (determinant/label):} Êthe linguistic label of a phrase is the complex phase of the matrix determinant.}Ê
\bigskip

The requirement of the ``complex phase"\footnote{For a complex number $z$, its norm is given by $|z| =  |\sqrt{z^* z}|$, with $z^*$ its complex conjugate. The complex phase is then defined as $z/|z|$, and quantifies the angle of $z$ in its representation in the complex plane.}, instead of taking directly the determinant, will become obvious when dealing with reasonable chain conditions in Sec.\ref{sec12b}. Thus, the Jarret-graph together with the previous axioms fixes the determinant of a matrix as the mathematical equivalent of a category label in linguistics. Let us be a bit more specific about this. The Jarret-graph codes the basic ``endocentric" nature of syntactic representations, provided that we make the following Auxiliary Assumption for the labels of the Chomsky Matrices:

\clearpage

\begin{figure}
	\centering
	\includegraphics[width=0.68\linewidth]{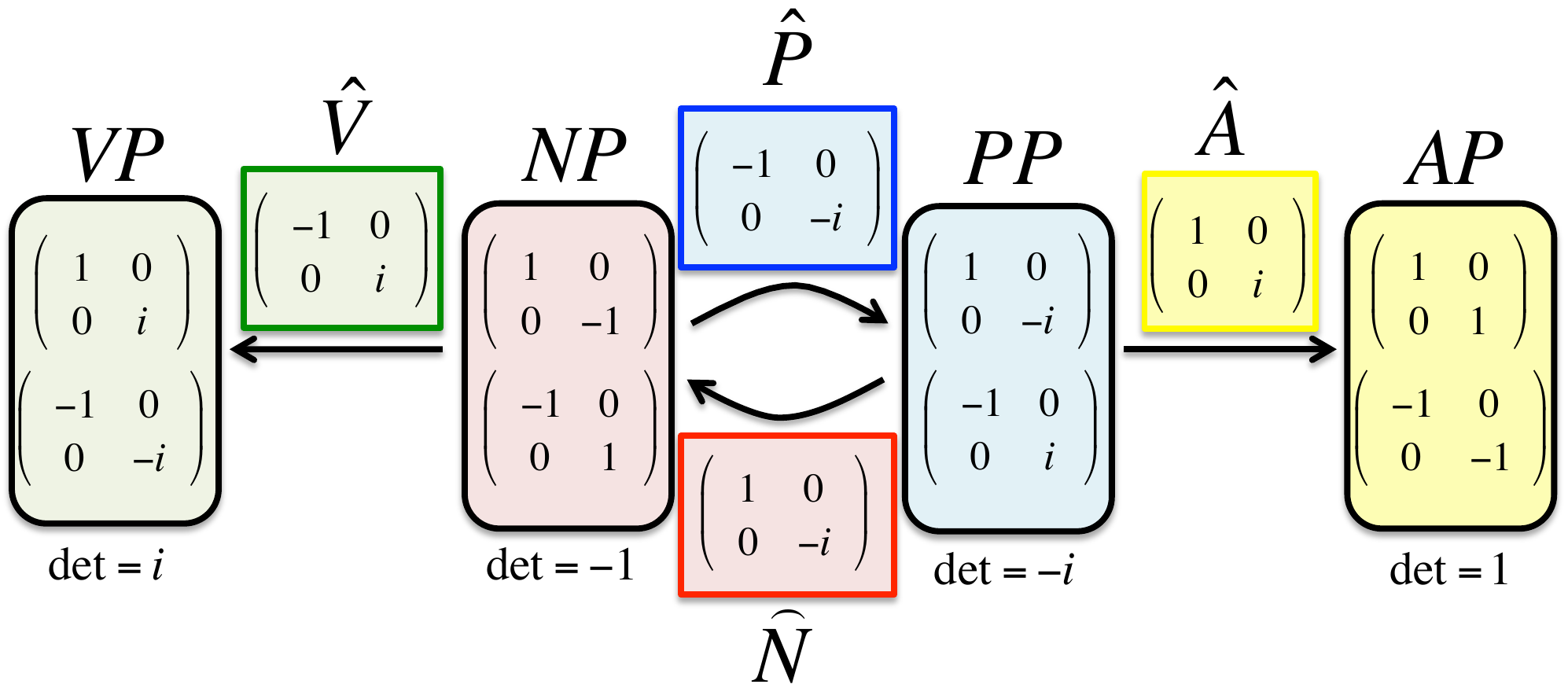}
	\caption{a valid representation of the Jarret-graph in matrix syntax. Colors are a guide to the eye. The determinant of the matrices corresponding to phrases is also shown.}
	\label{fig5}
\end{figure}

\bigskip
$\bullet$ \emph{{\bf Auxiliary Assumption:} Noun is lexically specified as $-1$, Verb is lexically specified as $i$, Adjective is lexically specified as $1$, and Elsewhere is lexically specified as $-i$.}
\bigskip 

Readers can see that, per the Auxiliary Assumption, the Jarret-graph simply instantiates the following broad Projection Theorem:

\bigskip
$\bullet$ \emph{{\bf Projection Theorem:} Lexically specified category $x$ projects to a phrase with label $x$,
for $x = \pm 1, \pm i$.}
\bigskip 

Given the structure of the matrices in the graph, the above is a theorem and not an axiom. Moreover, our Auxiliary Assumption above can also be understood as the logical consequence of the graph together with Axioms 1 to 7.  Finally, notice that the fact that all the matrices in Eq.(\ref{ax5}) have a \emph{cyclic structure} in the graph, which is  a consequence of them forming an abelian group, which will be analyzed in the next section.  

Before entering into the details of such group, a linguistic discussion is in order. Bear in mind that multiplying the matrices in Eq.(\ref{ax5}) among themselves results in matrices whose determinants are equivalent to the product of the factors' determinants. Moreover, it is easy to see that those results \emph{a fortiori} reduce to the orthogonal numbers Ê(in the real plane using the real inner product) \color{black} $\pm 1$ and $\pm i$. \emph{Ipso facto} that entails that the 64 possible multiplications arising in the group will present four sets of sixteen combinations, each of which has the same determinant/label. Because the group is commutative, those 64 combinations should reduce to 32 substantively (order being irrelevant). Moreover, we saw ``label duality" in instances in which it was not pernicious, since it corresponds to the ``twin" situations identically colored in the version of the Jarret-graph in Fig.\ref{fig5} --- thereby reducing the 32 substantively different options to 16 ``twin" instances. However, there also exists massive ambiguity in the products, since in the end these matrices form a group of 8 elements (as we shall see in the next section), which ``modulo an overall sign" can even be reduced to 4: ${\mathbb I}, Z, C_1$ and $C_2$. 

\begin{figure}
	\centering
	\includegraphics[width=0.45\linewidth]{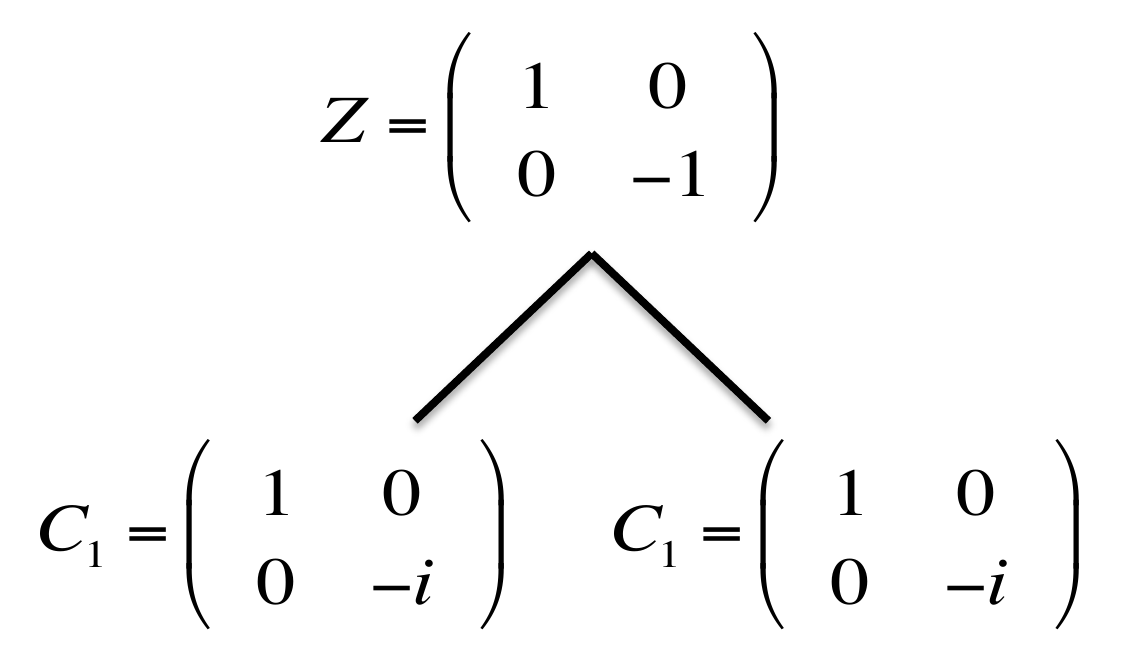}
	\caption{self-M of $C_1$ into $Z$.}
	\label{fig56}
\end{figure}

Coming briefly back to Axiom 6, we notice that its neat result is the projection in Fig.\ref{fig56}, which in turn ``freezes" $C_1$ as $N$ and $Z$ as $NP$. Any other use of $C_1$ as a head or $Z$ as a maximal projection is immediately prevented by the semiotic assumption. 

But there is more. We obtained $Z$ via self-M in Fig.\ref{fig56}, but we should be able to obtain the exact same mathematical results with $C_1$ being a projected category --- which still leads to $Z$ as a valid projection from the lexical $C_1$. And the way ``twin" categories are supposed to operate, we also know that $-Z$ has to be the ``twin" $NP$ projection, which therefore also means there must be a way for $C_1$ to take $-C_1$ as an argument so as to yield $-Z$ as its ``projected" value. So there is a projected $\pm Z$ category which we know is $NP$ in both ``twin" versions, and also a $\pm C_1$ of which we, therefore, know is not an $NP$ projection --- so it must be $VP$, $AP$, or $PP$.

We also started the entire enterprise based on ``conceptually orthogonal" distinctive features, which is why we chose a representation in terms of $\pm 1$ and $\pm i$ to start with --- as mathematically orthogonal entities in the complex plane. It may be taken as a further consequence of the semiotic assumption that whatever was orthogonal prior to projection remains orthogonal thereafter, or that orthogonality is preserved in projection. If so, once we start in Chomsky's orthogonal ``noun" and ``verb" categories understood as in Axiom 2, and we have determined $\pm Z$ to be the $NP$ projection, the $VP$ projection should be orthogonal vis-\`a-vis $\pm Z$. So given that $\pm Z$ has determinant/label $-1$, a possibility is that the orthogonal label is $i$ for $VP$, which is exactly what comes out from the graph and Axioms 1 to 7. The reason why $VP$ has label $i$ and not something else is, at this stage, a consequence of everything we said so far, and needs no further justification in itself.  

At that point the fate of all categories is sealed by the graph and Axioms 1 to 7. If $VP$ understood as $\pm C_2$ has to project from Chomsky's $-C_1$, the only factor that can thus multiply yielding the right result is $\pm Z$ --- so verbs must combine with $NP$s to yield proper projections. We leave it as an exercise for the reader to ascertain how the other combinations within the Jarret-graph follow from the same assumptions: once a possibility emerges, it rules out other equivalent ones (per the semiotic assumption). To be sure: had we started in a different syntactic anchor, these decisions would be radically different. 

It is worth emphasizing that the facts in Table \ref{tabHC} above are typically blamed on factors external to the language faculty, like general ``cognitive restrictions". It isn't obvious to us, however, what theory actually predicts most of those ungrammatical combinations, which were quite purposefully selected so as to ``make sense". Thus, for instance, to us an ungrammatical expression like the one in Table \ref{tabHC}(iv.c) --- namely, ``*in tall" --- presents a perfectly reasonable way to denote the state of being tall, which for example adolescents experience after a growth spurt. The issue is to understand why such a sound combination of concepts is ungrammatical, and this is precisely what the Projection Theorem prevents. 

\section{The ``magnificent eight" group $G_8 \simeq {\mathbb Z}_2 \times {\mathbb Z}_4 $}Ê
\label{sec8}

The eight matrices found in the representation of the Jarret-graph in Fig.\ref{fig5} have the mathematical structure of an \emph{abelian group}. We call this group $G_8$, or colloquially the ``magnificent eight" group. The elements in the group are  
\beq
G_8 = \{ \pm {\mathbb I}, \pm Z, \pm C_1, \pm C_2 \}, 
\eeq
with 
\beq
\mathbb{I} = \left(
\begin{array}{cc}
1 & 0 \\Ê
0 & 1 \end{array} \right), ~~
Z = \left(
\begin{array}{cc}
1 & 0 \\Ê
0 & -1 \end{array} \right), ~~
C_1 = \left(
\begin{array}{cc}
1 & 0 \\Ê
0 & -i \end{array} \right), ~~
C_2 = \left(
\begin{array}{cc}
1 & 0 \\Ê
0 & i \end{array} \right).
\eeq
The group is abelian (commutative) because all matrices are diagonal and therefore mutually commute. It is easy to check that this set satisfies the usual mathematical properties of a group: existence of a neutral element ${\mathbb I}$, an inverse for every element, and so forth. The group is also generated by the repeated multiplications of $C_1$, or of $C_2$. 

Ê
One last analytical point to bear in mind is that the group $G_8$ is in fact isomorphic to ${\mathbb Z}_2 \times {\mathbb Z}_4 $, i.e.,
\beq
 G_8 \simeq {\mathbb Z}_2 \times {\mathbb Z}_4 
 \eeq
The above equation means that $G_8$ and ${\mathbb Z}_2 \times {\mathbb Z}_4$ have, essentially, the same properties as far as groups are concerned. To see this more clearly, notice that the group $G_8$ includes several interesting subgroups, such as the $4$-element cyclic group 
\beq
G_4 = \{{\mathbb I}, Z, C_1, C_2\} \simeq {\mathbb Z}_4, 
\eeq
isomorphic to ${\mathbb Z}_4$ and with generator $C_1$ or $C_2$. Thus, by considering the semidirect product ${\mathbb Z}_2 \times {\mathbb Z}_4$ with ${\mathbb Z}_2 = \{ +1, -1 \}$, one obtains exactly a group that is isomorphic to $G_8$. 
\color{black} 

At this point, we choose to present the forthcoming derivations using the structure provided by this group, as this allows us to explain the basic features of the MS model. However, we would like to remark that the set of lexical entities that can be described by this group may be limited to the somewhat idealized circumstances alluded to in Fig.\ref{fig3}, where we purposely left out mention of so-called grammatical or functional categories. To address such a concern, in Sec.\ref{sec12} we introduce an extra ``symmetry" axiom that allows us to enlarge the relevant working group. Such an extension will be minimal, based only on symmetry considerations, and will have the aim of providing additional structure to the one obtained from $G_8$. In the meantime, we continue using group $G_8$ in our explanation, unless stated otherwise.

\clearpage 

\part{From groups to chains}
\label{p2}

\section{The Hilbert space $\mathcal{H}_8$: a linguistic qubit} 
\label{sec9}

We will see in Sec.\ref{sec10} that the formation of chains is modelled by sums of matrices. This entails the structure of a vector space, which will turn out to provide a natural solution to the problem of chain occurrences. In this section, we show how this vector space emerges in MS, and that it can be modelled by a Hilbert space. For this, we proceed step by step, starting from group $G_8$ as described in the previous section, including the sum, defining a scalar product, and analyzing the resulting structure.

\subsection{Dirac notation and scalar product} 

Before digging into more details, we need first to define and clarify a couple of concepts. The first one is that of a \emph{scalar product} between matrices. There are many options for this. Here we choose to work with the standard formula
\beq
\langle {A}|{B} \rangle = \tr(A^\dagger B).
\label{eq3}Ê
\eeq
For the sake of convenience, we will be using the Dirac (bra-ket) notation from quantum mechanics. Matrix $A$ is represented as $\ket{A}$, its adjoint $A^\dagger = (A^T)^*$ is represented by $\bra{A}$ ($T$ is the transpose, which  exchanges rows by columns, and $*$ is the complex conjugate, which substitutes $ i \leftrightarrow -i$), and the scalar product of $A$ and $B$ is repesented as $\langle {A}|{B} \rangle$. The scalar product produces a single complex number out of the two matrices. The product provides a way to define \emph{angles} between vectors in the space. For perpendicular (orthogonal) vectors, this scalar product is zero. 

We are going to build a vector space out of matrices, which can themselves be understood as elements of an abstract vector space\footnote{The fact that matrices, as such, can also form vector spaces by themselves, is well-known in linear algebra. See, for instance, the operator space in functional analysis: https://en.wikipedia.org/wiki/Operator$\_$space.}. Whenever a scalar product is zero, we will say that the corresponding matrices --- understood as abstract vectors --- are ``mutually orthogonal", in the sense that they are ``perpendicular"\footnote{Not to be confused with the concept of ``orthogonal matrix" in linear algebra, which is a matrix $A$ satisfying $A^T A = A A^T = {\mathbb I}$.}. For instance, one can easily check that ${\mathbb I}$ and $Z$ are mutually orthogonal, i.e., $\langle {{\mathbb I}}|{Z} \rangle = 0$. The \emph{norm} $N$ of a vector $\ket{A}$ (in our case, matrix $A$) is thus given by  
\beq
N = \left| \sqrt{\langle {A}|{A} \rangle} \right|Ê =  \left|Ê\sqrt{ \tr(A^\dagger A)} \right|. 
\eeq
A \emph{normalized} vector is a vector with norm $N=1$, for instance, 
\beq
\ket{\hat{A}} = \frac{1}{ \sqrt{\langle {A}|{A} \rangle}} \ket{A}, 
\eeq
where the ``hat" in this context means ``normalized", i.e., that the vector has norm one. 

Also, using these rules it is easy to see which pairs of matrices are mutually orthogonal when taking tensor products of the style $A \otimes B$, which as we explained amounts to elsewhere M in our model. One can then see that 
\beq
 \langle A \otimes B | C \otimes D \rangle = \tr(A^\dagger C) \tr(B^\dagger D) .  
\eeq
 This construction will be important in Sec.\ref{sec10}  when talking about chains. We will see that the linguistic idea of ``two occurrences of the same token in a derivation" amounts to vector sum in the space of the matrices obtained after elsewhere M (i.e., with a tensor product structure). It will be important that these matrices being summed correspond to orthogonal vectors. ÊMathematically speaking, our scalar product defines an inner product between elements in the space, and therefore this is a normed vector space. If we do not restrict the coefficients in the sums of elements in this space,  then the vector space is actually a complex Hilbert space, i.e., a complete complex vector space on which there is an inner (scalar) product associating a complex number to each pair of elements in the space. However, we will see in the forthcoming sections that, \emph{in practice}, one does not need arbitrary superpositons if inguistic chains are the only object to be concerned with. At this level, thus, our space does not need to be a complete space (i.e., it has ``holes"), and is what is called a pre-Hilbert space. We believe, however, that a complete Hilbert space should be useful in future developments, apart from being much more natural (at least from a physicist's point of view). But it is important to keep in mind that, perhaps, language may be describable by a vector space that is not complete. \color{black}ÊAll in all, in the next subsection we show that there is a very natural $2d$ space emerging from everything that we have said so far.  

\subsection{From group $G_8$ to space $\mathcal{H}_8$.}

Let us now see which pairs of matrices in group $G_8$ are mutually orthogonal. The scalar groups of the different matrices in $G_8$ can be obtained from those in Table \ref{tab1}. Rows in this table corresponds to the matrix for which there is a $\dagger$ in the scalar product formula of Eq.(\ref{eq3}) (the "bra"), and columns for those that do not (the ``ket"). Products between other elements in $G_8$ are also given by the numbers in the table, up to possible $\pm 1$ and $\pm i$ multiplying factors. 

\begin{table}[h]
\centering
\begin{tabular}{ ||c || c | c | c | c ||}
\hline
 & ${\mathbb I}$  & $Z$ & $C_1$ & $C_2$  \\ \hline \hline 
${\mathbb I}$ &  $2$ & $0$ & $1-i$ & $1+i$  \\\hline 
$Z$ &         $0$      &   $ 2$ &    $ 1+i$ &  $ 1-i$ \\\hline 
$C_1$&   $1+i$    & $ 1-i$ & $ 2$ & $ 0 $   \\ \hline
$C_2$ &   $ 1-i$ &   $1+i$ &     $ 0 $     &     $  2 $  \\ \hline 
\end{tabular}
\caption{Table of scalar products of some elements in the $G_8$.}
\label{tab1}
\end{table}

Table \ref{tab1} provides the dimensionality as well as possible choices of basis of a vector space constructed by the elements in $G_8$.  The space generated by linear combinations of matrices in $G_8$ is the $\mathcal{H}_8$ Hilbert space, i.e., a space where the matrices of ${G}_8$ are understood as abstract vectors, and which can be superposed being multiplied by arbitrary complex coefficients. \color{black}ÊThis structure means that vectors such as this one are allowed to exist: 
\beq
\ket{\psi} = \alpha \ket{{\mathbb I}} + \beta \ket{Z} + \gamma \ket{C_1} + \cdots
\eeq
In the above equation we adopted again the Dirac notation for vector $\ket{\psi} \in \mathcal{H}_8$ and the elements in the sum. The coefficients $\alpha$, $\beta$ and $\gamma$ are complex numbers, and the elements in the superposition are, as noted, matrices from group ${G}_8$ understood as (not normalized) vectors from $\mathcal{H}_8$. The resulting vector $\ket{\psi}$, understood as a matrix, is not necessarily an element of  ${G}_8$, but it is a valid vector of the Hilbert space $\mathcal{H}_8$. As an example, the vector 
\beq
\ket{\psi} = \ket{{\mathbb I}} + \ket{Z}
\eeq
is a perfectly valid vector in $\mathcal{H}_8$, but its associated matrix
\beq
{\mathbb I}Ê+ Z = 
\left(
\begin{array}{cc}
2 & 0 \\Ê
0 & 0 \end{array} \right)
\eeq
is not an element of the group $G_8$. 

Consider next the dimension of $\mathcal{H}_8$, which is the number of linearly-independent matrices in ${G}_8$\footnote{As a reminder, $n$ matrices $A_i$, with $i = 1, \ldots, n$, are linearly independent if there are no complex numbers $\alpha_i$ such that $\alpha_1 A_1 + \alpha_2 A_2 + \cdots + \alpha_n A_n= 0$.}. This can be obtained from the maximum number of mutually-orthogonal matrices in Table \ref{tab1}. We see, thus, that in $G_8$ there are always at most 2 of such matrices, for instance the Pauli matrix $Z$ together with the identity
\beq
\{  {\mathbb I}, ~ Z  \},  
\eeq
but also the two Chomsky matrices
\beq
\{ C_1,  ~ C_2 \}. 
\eeq
To see this, notice that the two Chomsky matrices above can in fact be written in terms of the Pauli matrix $Z$ and the identity as follows: 
\beq
C_1 = \left( \frac{1-i}{2} \right) {\mathbb I} + \left( \frac{1+i}{2} \right) Z ,~~~~~~~~ C_2 = \left( \frac{1+i}{2} \right) {\mathbb I} + \left( \frac{1-i}{2} \right) Z. 
\label{change11}
\eeq
These relations can also be inverted easily, in order to find the Pauli matrix $Z$ and the identity from the two Chomskyan matrices: 
\beq
{\mathbb I} = \left( \frac{1+i}{2} \right) C_1 + \left( \frac{1-i}{2} \right) C_2 ,~~~~~~~~ Z = \left( \frac{1-i}{2} \right) C_1 + \left( \frac{1+i}{2} \right) C_2.
\label{change12}
\eeq
Therefore, the dimension of the Hilbert space $\mathcal{H}_8$ associated to $G_8$ is 2, i.e., 
\beq
{\rm dim}(\mathcal{H}_8) = 2. 
\eeq
That is an interesting result, since starting with the simplest set of axioms, we have arrived at the simplest non-trivial Hilbert space: one of dimension 2. This space is isomorphic to the Hilbert space of a qubit, i.e., a 2-level quantum system, namely, 
\beq
\mathcal{H}_8 \simeq {\mathbb C}^2.  
\eeq
Our conclusion is that the straightforward linguistic Hilbert space $\mathcal{H}_8$, obtained from $G_8$, is in fact \emph{the linguistic version of a qubit}\footnote{For those unfamiliar with quantum computation, a qubit means a ``quantum bit", and amounts to a quantum 2-level system. The description of such a system is done with a $2d$ Hilbert space called ${\mathbb C}^2$. An arbitrary state $\ket{\psi}$ in this Hilbert state is usually written as $\ket{\psi}  = \alpha \ket{0} + \beta \ket{1}$, with $\alpha, \beta$ complex numbers such that $|\alpha|^2 + |\beta|^2 = 1$ (normalization), and $\{Ê\ket{0}, \ket{1} \}$ some orthonormal basis. Contrary to the notion of classical bit, which can only have values $0$ or $1$, a qubit can have values $\ket{0}$, $\ket{1}$, and any superposition thereof. Typical physical examples include the angular momentum of a spin-$1/2$ particle (spin up/down), superconducting currents (current left/right), atomic systems (atom excited or not), and many more. See, e.g., Ref.\cite{NielsenChuang} for more details.}.  
 
In order to characterize this vector space, next we find an orthonormal basis, which is the minimal set of orthogonal vectors with norm one ($=$ orthonormal), such that any other vector in the space can be written as a linear combination of these. From Table \ref{tab1} one can see that a possible orthonormal basis for the space $\mathcal{H}_8$ is given by the normalized Pauli $Z$ matrix together with the normalized identity:
\beq
{\rm  Pauli ~ basis} \equiv \left\{  \frac{{\mathbb I}}{\sqrt{2}},   ~ \frac{Z}{\sqrt{2}}  \right\} \equiv  \left\{ \ket{\hat{{\mathbb I}}},  ~\ket{\hat{Z}} \right\}, 
\eeq
where in the last step we used again the Dirac notation. What this means is that any vector in $\ket{\psi}  \in \mathcal{H}_8$ can be written as 
\beq
\ket{\psi} = \alpha \ket{\hat{I}}  + \beta \ket{\hat{Z}}, 
\eeq
where $\alpha$ and $\beta$ are complex numbers ---  ``coordinates" of vector $\ket{\psi}$ in this Pauli basis.  This basis is quite appealing since all the vectors on it correspond, themselves, to Hermitian matrices. Moreover, it also follows from all the above that there is yet another natural basis for $\mathcal{H}_8$ given by ``normalized" Chomsky matrices: 
\beq
{\rm  Chomsky ~ basis} \equiv \left\{  \frac{C_1}{\sqrt{2}},  ~ \frac{C_2}{\sqrt{2}}  \right\} \equiv  \left\{ \ket{\hat{C_1}}, ~\ket{\hat{C_2}} \right\}. 
\eeq
Any vector in $\mathcal{H}_8$ can be written in that valid basis as 
\beq
\ket{\psi} = \alpha' \ket{\hat{C_1}}  + \beta' \ket{\hat{C_2}}, 
\eeq
where $\alpha'$ and $\beta'$ are the ``coordinates" of vector $\ket{\psi}$ in this Chomsky basis. Indeed, the relation between these two basis is simply a unitary matrix, which according to Eq.(\ref{change11}) and Eq.(\ref{change12}) can be represented like this: 
\beq
\begin{pmatrix}
  \alpha \\
  \beta 
 \end{pmatrix} = \begin{pmatrix}
\frac{1-i}{2} & \frac{1+i}{2}  \\Ê
\frac{1+i}{2} & \frac{1-i}{2}  
\end{pmatrix} \begin{pmatrix}
  \alpha' \\
  \beta' 
 \end{pmatrix}.
 \eeq
The above matrix-vector multiplication gives us the rule to change from ``Chomsky coordinates" to ``Pauli coordinates" for any vector in the ``linguistic qubit" Hilbert space $\mathcal{H}_8$. Both ``coordinate systems" are valid, as far as the description of the space is concerned. It is also interesting to note that both bases include 2 out of the 4 original lexical categories as labeled by the matrix determinant. In the convention used above, for example, the Pauli basis includes determinants $\pm 1$, i.e., $NP$ and $AP$, whereas the Chomsky basis includes determinants $\pm i$, i.e., $VP$ and $PP$. 

\section{A model for linguistic chains}
\label{sec10} 

Before entering into the details of what chains are and how we can model them with the mathematical machinery we have presented, letÕs review the difference between external M (EM) and internal M (IM), which will be important in understanding chains. So far, we have mostly been concerned with EM, an operation between two independent linguistic elements in a derivation (except in the special case of self-merge discussed above, which is also EM, but involving only one element), either of which may in principle come directly from the lexicon (hence a derivational atom) or constitute the result of a previous series of M operations (hence a complex phrase/projection of some sort). IM is a somewhat less intuitive process in which one merges an entire phrase with \emph{one of its constituent parts}. That is to say: an element $L$ that is contained in some element $K$, which has already been assembled by M, somehow ``reappears" and merges with $K$. 

To see better how that can happen, consider again the example in Sen.\ref{ls1} and repeated below as in Sen.\ref{ls43}, here together with a bracket analysis:
\beqa
\label{ls43} 
&a.& {\rm Alice ~ seems ~to ~ like ~Bob.} \\ 
&b.& [_{TP} ~{\rm Alice}_k ~ [_{T'} ~TENSE ~ [_{VP}Ê~ {\rm seems} ~ [_{TP} ~ {\rm Alice}_k  ~[_{T'} ~ [_T ~{\rm to}~ ] ~ [_{VP} ~ {\rm like ~ Bob} ~]_{VP} ~ ]_{T'} ~ ]_{TP} ~ ]_{VP} ~]_{T'} ~]_{TP} \nonumber 
\eeqa
In this sentence, the element \emph{Alice} has moved from its original position (subject of the $VP$ \emph{like Bob}) to the beginning of the sentence, creating multiple occurrences of \emph{Alice}, one in the original configuration and one as the newly created specifier. In terms of M, such a displacement is modelled by having the higher $T'$ (namely, the immediate projection of $TENSE$, which is the root structure at this point in the derivation, prior to creation of the specifier position) combine with the occurrence of \emph{Alice} in the specifier of \emph{to}, which results in a subsequent projection, $TP$. Since Alice is contained in the higher $T'$, this merger operation is IM. 

As we have noted, one of our main motivations in the present project is precisely to understand what it means for such occurrences to be a way for an item like \emph{Alice} to be distributed over a phrase marker by way of IM along the lines in Sen.\ref{ls43}(b), after internal M (IM). Armed with the tools in previous sections, we are now in a position to describe linguistic chains\footnote{ÊSince the present paper is focusing on mathematical foundations, we will not go into various linguistic issues that arise, as discussed in our forthcoming monograph. For example, we focus on ``subject-to-subject" displacement first, not ``object-to-subject" conditions (as in passives -- though see Sec.\ref{sec131}). We are also not emphasizing now recursive conditions that arise with multiple specifiers, or how selection conditions could change when a phrase does or does not take a specifier, and many others. While these are important analytical situations to expand on, they do not change the mathematics of the situation. What we are discussing here, after having created an actual group under natural linguistic circumstances, involves off-the-shelf mathematics, for which in particular configurational collapses as we are about to review in context are necessary for tensorized dependencies of orthogonal categories. That is the main problem we are trying to solve (the distribution and interpretation of syntactic occurrences). The entire syntactic exercise, therefore, does not attempt to be comprehensive, but is merely presenting a teaser of what we deploy in the monograph.  \color{black}Ê}. We will take movement chains as in Sen.\ref{ls43} to arise from instances of EM of the \emph{elsewhere} (i.e., not 1st M) sort --- thus requiring tensor products. In linguistic jargon, chains are $\{ \{ \alpha, K \}, \{ \alpha, \Lambda \} \}$ objects, where $\alpha$ (e.g. \emph{Alice} in Sen.\ref{ls43}) involves (prior) context $\Lambda$ and (subsequent) context $K$ as separate occurrences, which in a derivational sense go \emph{from $\Lambda$ to $K$}. Note that we need not code the fact that $\Lambda$ is prior to $K$ simply because \emph{$K$  properly includes $\Lambda$} --- this is easy to see in Sen.\ref{ls43}, where the higher $T'$ ``dominates" the lower $T'$ (an entirely different token projection of the same type). We have highlighted both projections so that this simple point is clear. 

In Dirac notation, what we have said corresponds to the vector
\beq
\ket{{\rm Chain}} = \frac{1}{\sqrt{2}} \left(\ket{\hat{\alpha}} \otimes  \ket{\hat{K}}
 + \ket{\hat{\alpha}} \otimes  \ket{\hat{\Lambda}} \right) =  \frac{1}{\sqrt{2}} \ket{\hat{\alpha}} \otimes \left( \ket{\hat{K}} + \ket{\hat{\Lambda}} \right), 
 \label{chain1}
 \eeq
 where we assume normalized vectors everywhere, and also assume --- so far merely for convenience ---  that the resulting vector must be normalized to 1\footnote{ÊIn fact, the bilinearity in Eq.(\ref{chain1}) is a consequence of the properties of the tensor product. \color{black}}. The above equation may be clearer for a physicist than for a linguist. An alternative bracketed notation, perhaps more linguist-friendly, would be
\beq
[ {\rm Chain} ] = [ \alpha  \otimes  K] 
 + [\alpha \otimes \Lambda] =  \alpha \otimes [K + \Lambda].  
 \eeq
In a valid chain, a specifier is at two different positions. In this example, $\ket{\hat{\alpha}}$ acts as the specifier of $\ket{\hat{K}}$ and $\ket{\hat{\Lambda}}$ \emph{simultaneously}. One would then say that the specifier is ``at two positions at the same time". In the MS model this is achieved through superposition of the elements that are being ``specified", which in this case amounts to $\left( \ket{\hat{K}} + \ket{\hat{\Lambda}} \right) $. In the physics jargon, one would say that a linguistic chain is a superposition of two different states, both of them specified by the same linguistic specifier. To put it in popular terms: within the context of MS, a chain is a linguistic version of Schr\"odinger's cat, as it is based on the equally-weighted superposition of two (orthogonal) vectors in our (linguistic) Hilbert space.  

We elevate this construction to the category of axiom: 

\bigskip
$\bullet$ \emph{{\bf Axiom 8 (Hilbert space):}Ê linguistic chains are normalized sums of vectors from a Hilbert space.}Ê
\bigskip

One side remark is in order: in the discussion above we have only considered what linguists call A-chains. These arise in the sorts of ``raising" constructions reviewed in this section (as well as passives and other related constructions). There are many other instances of displacement,  for instance:
\beqa
\label{ls46}Ê
&a.& {\rm Beans, ~Êmy ~Êdog ~ likes.} \\Ê
&b.& {\rm What ~Êdoes ~Êyour ~ dog ~like?} \nonumber \\Ê
&c.& {\rm These ~are ~the ~beans ~that ~your ~dog ~likes.} \nonumber \\Ê
&d.& {\rm Beans ~are ~not ~so ~hard ~for ~dogs ~to ~like.} \nonumber
\eeqa
These are commonly called A'-chains, which for various reasons are significantly more complex. We have a well-known way to deal with these more elaborate constructions also, but to keep the discussion focused we will not go into it now\footnote{Let us say at least that such A'-chains may involve \emph{entangled} states in the Hilbert space, i.e., states that cannot be separated as tensor products of individual vectors. As we will see in the next section, this is not the case of A-chains, which are separable (non-entangled) states.}. 

We can now push the mathematical machinery further: when the (A) chain is sent to the phonetic and semantic interfaces (i.e., is \emph{spelled out} and \emph{interpreted}), only one of the two valid configurations for the chain occurrences is realized. A priori, one does not know which one of the two is realized, since both are valid options. We note that this process, in the context of MS, has a close mathematical analogy with the measurement postulate of quantum mechanics: once a superposed quantum system is observed in the appropriate basis, it ``collapses" in one of the superposed options as a result of the measurement, which also dictates the measurement outcome. Within the mathematical context that we propose, this is indeed also a logical option, at least mathematically, for the case of chains at an interface. 

We elevate that to the category of an axiom: 

\bigskip
$\bullet$ \emph{{\bf Axiom 9 (interface):}Ê when a chain $\ket{\psi}$ is sent to an interface, its vector gets projected in one of the elements $\ket{\phi}$ being specified according to the probability $|\braket{\phi}{\psi}|^2$.}Ê    
\bigskip

For instance, in the example of Sen.\ref{ls43}, the chain has a 50$\%$ probability of configurationally collapsing in either context $K$ or context $\Lambda$ (should these be orthogonal), when sent to the interfaces. As discussed in Sec.\ref{sec2}, this is  the puzzling behavior of chains, whose occurrences are somehow distributed through the configurations they span over as some key element ``displaces" via IM, in terms of the information such objects carry. For us the reason why the discontinuous element cannot materialize in all the configurations it involves, or a few of them, or none of them (but must collapse in precisely one) has a close mathematical analogy with the reason why the angular momentum of a spin-$1/2$ particle (e.g., an electron) can exist simultaneously in ``up" and ``down" configurations of its $z$-component, but is only observed at one of them when being measured. 

To conclude this section, we note that there are superpositions in the Hilbert space we have been considering that do not correspond to grammatical objects. In other words, there is more structure available in our construction than what is empirically needed. However, one can filter ``non-grammatical" vectors by means of extra conditions, whether mathematical or linguistic in some sense (broadly cognitive, semiotic, information-theoretic, etc.). We discuss this in the next section, together with other problems intrinsically related to chains.

\section{Filtering and compression}
\label{sec11} 

Having presented our approach to linguistic chains, in this section we present some issues with the current description that can be elegantly solved. The first is the issue of the great abundance of states in the system, including possible states that seem ungrammatical on empirical grounds (constructions corresponding to sentences that speakers deem unacceptable). The second is the explosion of the relevant matrix size when doing tensor products according to elsewhere M.

\subsection{Valid grammatical derivations and separable states}Ê
 
The Hilbert space describing a chain includes valid grammatical vectors, as we have seen, but also many others that may be regarded as ungrammatical. For instance, consider the ``weird" vector
\beq
\ket{{\rm Weird }} = \frac{1}{\sqrt{2}} \left(a \ket{\hat{\alpha}} \otimes  \ket{\hat{K}}
 + b \ket{\hat{\beta}} \otimes  \ket{\hat{\Lambda}} \right), 
 \eeq
 or in bracketed linguistic notation 
 \beq
[ {\rm Weird} ] = a [ \alpha  \otimes  K] 
 + b [\beta \otimes \Lambda].  
 \eeq
For $a, b \neq 0$, this state is what we call \emph{entangled}, using the quantum information jargon; that is, the state cannot be separated as a [``left vector" $\otimes$ ``right vector"]. This is also a valid state in our vector space. The point is, however, that at least up to our current knowledge, such a state may not be grammatical in any general sense: in does not appear to correspond to the result of any derivation according to syntax as presently understood. To the contrary, this looks like the superposition of two possible different, even unrelated, derivations, one with probability $|a|^2$ and the other with probability $|b|^2$. What sort of a formal object is that?

While it may seem tempting to speculate that such ``wild" comparisons do happen as we think to ourselves in ``stream of consciousness" fashion, or randomly walk into a party where multiple conversations are taking place, only bits and pieces of which we happen to parse --- we will not go there. Future explorations may well need to visit such territories, but we have nothing meaningful to offer now, in those terms. For starters, all our analysis happens within standard minimalist derivations, which range over possibilities that stem from a given ``lexical array". For example, when studying Sen.\ref{ls43} above, our lexical choices are restricted to \emph{Bob , Alice, like, seem}, and so on. In those terms it is meaningful to ask whether \emph{Bob} has raised or not, and if it has,  whether it is pronounced or interpreted at one position or another, etc. But it would not be meaningful to even ask how all of that relates to a derivation about \emph{John, Mary, hate}, or whatever. It may be that such a thought is somehow invoked in a speaker's mind when speaking of Bob and Alice --- but current linguistics has literally nothing to offer about that yet. 

This is not to say that the issue is preposterous, in any way. Psycholinguists have known for decades, for example, that (in some sense) related words ``prime" for others, enhancing speakersÕ speeds at word-recognition tasks (e.g., \emph{hospital} primes for \emph{doctor} or \emph{top} for \emph{stop}) \cite{psycho}. If what we are modelling as a Hilbert space is a reflex of the human lexicon ultimately being a Hilbert space --- as proposed by Bruza \emph{et al.} in Ref.\cite{mathphys} --- then it would stand to reason that such matters could well be the reflex of ``lexical distance" metrics, in a rigorous sense. In any case, this is not the focus of our presentation, so we put that aspect of it to the side.

The presupposition of a ``lexical array" that given derivations are built from, certainly acts as a ``filtering" device into what lexical constructions even ``count" as objects of inquiry for our purposes. In what follows we consider other such ``filtering" notions, within the confines of narrowly defined derivations, in the broad sense of the MP. 

Within the scope of our MS model, states without chain conditions are of the kind  
\beq
\ket{{\rm Regular }} =  \ket{\hat{\alpha}} \otimes  \ket{\hat{K}},
\label{reg}
\eeq
and states of the kind 
\beq
\ket{{\rm Chain }} = \frac{1}{\sqrt{2}} \left(a \ket{\hat{\alpha}} \otimes  \ket{\hat{K}}
 + b \ket{\hat{\alpha}} \otimes  \ket{\hat{\Lambda}} \right) = \frac{1}{\sqrt{2}} \ket{\hat{\alpha}} \otimes \left( a \ket{\hat{K}} + b \ket{\hat{\Lambda}} \right), 
 \eeq
correspond to a chain --- where we allowed for different possible coefficients $a$ and $b$ (shall this ever be needed for whatever reason).  In the context of our construction, mathematically these are separable states: those states that have no entanglement in the ``quantum mechanical" jargon. Therefore it looks (for now anyway) as if only such states are grammatically correct. 

We elevate that to the category of an axiom for chains: 

\bigskip
$\bullet$ \emph{{\bf Axiom 10 (filtering):}Ê valid grammatical chains are separable (non-entangled) states in the Hilbert space.}Ê  
\bigskip

Note that Axiom 10 applies also when there are no chain conditions at all, i.e., ``trivial chains" leading to a regular derivation as in Eq.(\ref{reg}). Moreover, the Axiom does not say that all separable states are grammatically valid --- rather, that all gramatically-valid chains are separable states. This also implies that the relevant, grammatical, ``corner" of our linguistic Hilbert space corresponds, at least for chains, to the set of separable states. So if entanglement is a resource (as is advocated by quantum information theorists), it looks as if our language system naturally chooses to work in this case \emph{with the cheapest set of states}: those that have no entanglement at all, and which are known to be computationally tractable in all aspects. Even if this analogy to quantum mechanics is just at the level of the formalism, we find it intriguing in that it connects to the minimalist idea that language is some kind of optimal system, in the sense that it does not consume more computational resources than what it strictly needs to achieve its goals (what we described above as the 3rd factor of language design). Apparently, nature seems to prefer states with low entanglement, which may be true also for language in some cases\footnote{Let us mention that, beyond the conditions analyzed in this paper, it may be possible to find more complex valid grammatical structures that are accounted for by \emph{entangled} states in some Hilbert space. Examples are A'-chains, control, parasitic gaps, and ellipsis, just to mention a few cases (inasmuch as some of those operations also involve Agree, see Sec.\ref{newsec}). It will be interesting to understand mathematically which \emph{types} of entanglements, if any, are found in such more complicated derivations, as well as the \emph{amount} of such entanglements. As said, we leave this exploration for future works.}.  

 \subsection{Explosion and compression} 

Let's now come back to \emph{external} M, in its \emph{elsewhere} variety, which we have modelled via tensor products. For a recursive structure, every external M corresponding to recursive specifiers, even without chain conditions, will yield a new tensor product. We therefore have situations like:
\beqa
{\rm John's ~ car} ~ &\rightarrow& A \otimes W \nonumber \\Ê
{\rm John's ~ mother's ~ car} ~ &\rightarrow& A \otimes B \otimes W \nonumber \\Ê
{\rm John's ~ mother's ~ sister's ~ car} ~ &\rightarrow& A \otimes B \otimes C \otimes W \nonumber \\Ê
&\cdots&
\label{conc}
\eeqa
with $A,B,C$ and $W$ some of the $2 \times 2$ matrices in our system. The amount of information added to the system increases linearly with the number of specifiers, since this is proportional to the number of matrices being tensorized. However, the size of the actual matrix describing the derivation grows exponentially, and for a derivation with $n$ elements, this will be $2^n \times 2^n$. This is a computational blow up.

Based on the assumption that all of the MP happens within the broad domain of the computational theory of mind, it seems plausible to expect this kind of explosion to be somehow truncated as the syntactic derivation unfolds. 

We have good reason to believe that this information is encoded in the eigenspaces of the overall matrix. In the initial conditions of our system, determined by 1st M, it is easy to see that regular derivations without chain conditions can only have eigenvalues $\pm 1$ and $\pm i$, perhaps with degeneracies, implying that the number of such eigenspaces is always upper-bounded by 4 and, therefore, never explodes. For chains, however, the sum involved in chain formation (per Axiom 8) may lead to a blowing number of eigenspaces, as things get tensorized. To control the growth one therefore needs a truncation or filtering scheme that targets only the relevant eigenspace. The option we propose here is \emph{matrix compression}. More specifically, the matrix resulting from the construction of a chain must have some null eigenspace (i.e., some zero eigenvalues), which is therefore irrelevant for its description.

We elevate this to the category of a postulate: 

\bigskip
$\bullet$ \emph{{\bf Axiom 11 (compression):}Ê matrices of valid grammatical chains have a non-trivial null (irrelevant) eigenspace.}Ê  Ê
\bigskip

This axiom may look cumbersome, as we do not provide here any details about this null eigenspace, or how many null eigenvalues one should expect. In any case, the point of this axiom is that the explosion problem must be avoided when constructing valid chains, as otherwise they are not computationally efficient (and one could never deal with chains entailing representations as in Sen.\ref{conc}). Our claim is that we can do this by controlling the growth of non-trivial eigenspaces when doing tensor products (elsewhere Ms) combined with vector sums (chain formations). More details on this are provided in Secs.\ref{sec12b} and \ref{newsec}. 

\section{More structure via reflection symmetry}  
\label{sec12} 

In the discussion above, we have consistently ignored all the ``grammatical" or ``functional" elements that make up language. Just to give a sense of how much that is, observe:
\beqa
\label{ls52}Ê
 &a.& {\rm {\bf We ~Êhave}  ~Êconsistent{\bf ly} ~Êignore{\bf d ~Êall ~Êthe} ~Ê``grammatical" ~ element{\bf s}.} \\Ê
 &b.&  {\rm   ~~~~~~ ... ~~~~     ~     consistent   ... ~~ ignore      ...  ~  ...  ~~  ...  ~~  ~  ``grammatical" ~element... }Ê\nonumber \\Ê
 &c.& {\rm {\bf We ~ have  ~~~~  ~  ...    ~~~     -ly    ~~~       ...-d ~ all ~ the ~   ~~~~~~~~~    ...    ~~~~~~~~~ ~~~~ ...               -s}} \nonumber 
\eeqa
Our theory so far has concentrated on the ``tier" in Sen.\ref{ls52}(b), not the one in Sen.\ref{ls52}(c) --- but both such ``tiers" are known to be central in making up the representation in Sen.\ref{ls52}(a). 

In order to capture ``grammatical" elements, we need a larger group of matrices that also satisfies similar requirements as $G_8$ and which allows us to enlarge our ``periodic table" of language categories from those as in Sen.\ref{ls52}(b) to those as in Sen.\ref{ls52}(c) also. Notice also that the Hilbert space constructed from $G_8$ has dimension $2$ and is isomorphic to the Hilbert space of a qubit. While this is appealing, plausibly relating to the substantive information content of sentences (which is what the ``lexical tier" in Sen.\ref{ls52}(b) is about, as is apparent in ``telegraphic speech"), the larger Hilbert space is desirable when thinking of the larger picture of natural language syntax. The minimal addition in the system that allows us to build the simplest, non-trivial, extra structure is the assumption that if a diagonal matrix is allowed in the system with some elements along the main diagonal, then another matrix with the same elements, \emph{but along the antidiagonal}, should also be allowed. In other words, if a matrix
\beq C = \left(
\begin{array}{cc} 
a & 0 \\Ê
0 & b \end{array} \right)
\eeq
forms part of the system, then we also must allow 
\beq S = \left( 
\begin{array}{cc} 
0 & a \\Ê
b & 0 \end{array} \right)
\eeq
to be a matrix in the system as well. We call this property \emph{reflection symmetry}, and elevate it to the category of one of our axioms: 

\bigskip
$\bullet$ \emph{{\bf Axiom 12 (structure):} Êthe system has reflection symmetry.}
\bigskip

This axiom may seem artificial at first, but it allows us to build a coherent set of matrices that incorporates all the features discussed until now, as well as many features discussed by other authors in other papers --- which we will go to in the monograph \cite{mono}. It is also the minimal assumption to extend the set of matrices in MS in a non-trivial way, specially to a non-abelian group.

\subsection{The Chomsky-Pauli group $G_{cp}$} 

It is easy to see that the combination of Axiom 1 (Chomsky matrices), Axiom 3 (matrix multiplication as 1st M) and Axiom 12 (reflection symmetry), yields a unique set of 32 matrices with the structure of a non-abelian group. We call this the Chomsky-Pauli group $G_{cp}$, and it is given by 
\beq
G_{cp} = \{ \pm {\mathbb I}, \pm X, \pm Y, \pm Z, \pm i {\mathbb I}, \pm iX, \pm iY, \pm iZ, \pm C_1, \pm C_2, \pm S_1, \pm S_2, \pm iC_1, \pm iC_2, \pm iS_1, \pm iS_2 \}, 
\eeq
where ${\mathbb I}$ is the $2 \times 2$ identity matrix, $X,Y$ and $Z$ are the $2 \times 2$ Pauli matrices, $C_1$ and $C_2$ are two of the original ``Chomsky" matrices, and $S_1$ and $S_2$ are ``anti-Chomsky" (just to fix some notation) matrices. Explicitly, these 8 matrices are the following:  
\begin{eqnarray}
\mathbb{I} = \left(
\begin{array}{cc}
1 & 0 \\Ê
0 & 1 \end{array} \right), ~~
X = \left(
\begin{array}{cc}
0 & 1 \\Ê
1 & 0 \end{array} \right), ~~
Y = \left(
\begin{array}{cc}
0 & -i \\Ê
i & 0 \end{array} \right), ~~
Z = \left(
\begin{array}{cc}
1 & 0 \\Ê
0 & -1 \end{array} \right), \nonumber \\  
C_1 = \left(
\begin{array}{cc}
1 & 0 \\Ê
0 & -i \end{array} \right), ~~
C_2 = \left(
\begin{array}{cc}
1 & 0 \\Ê
0 & i \end{array} \right), ~~
S_1 = \left(
\begin{array}{cc}
0 & 1 \\Ê
-i & 0 \end{array} \right), ~~
S_2 = \left(
\begin{array}{cc}
0 & 1 \\Ê
i & 0 \end{array} \right).
\end{eqnarray}

\begin{figure}
	\centering
	\includegraphics[width=0.68\linewidth]{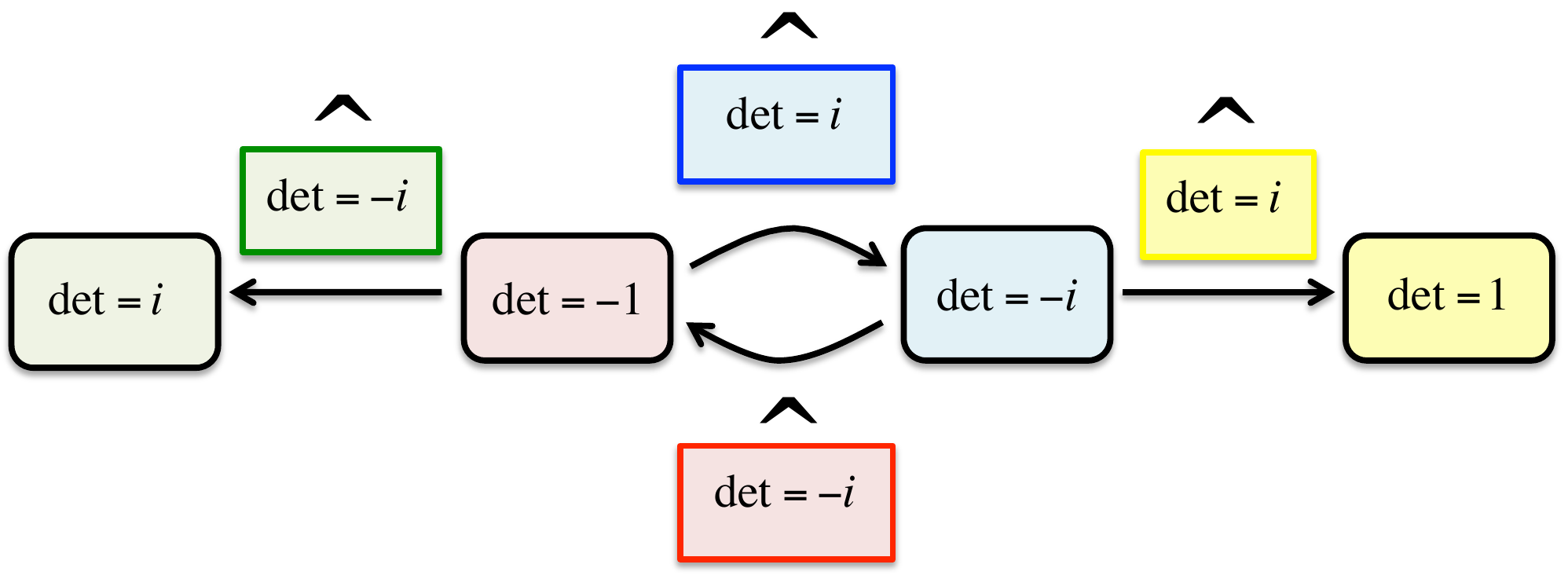}
	\caption{Jarret-graph in terms of determinants/labels for all instances. Colors are a guide to the eye.}
	\label{fig67}
\end{figure}

Several remarks are in order. First, we call this group the Chomsky-Pauli group for obvious reasons. It contains the original Chomsky matrices in Eq.(\ref{ax2}), and it also includes the Pauli matrices $X$, $Y$ and $Z$. In fact, the new group includes the Pauli group as a subgroup, as we explain in what follows. Second, the determinants of all the 32 matrices are either $\pm 1$ or $\pm i$, matching the fact that there are four possible lexical categories indexed by their labels, computed as determinants according to Axiom 7. This is important because ``grammatical" elements are known to strictly correspond to the ``substantive" categories --- which we can easily implement by keeping track of their respective determinant/labels. Third, all new categories in the extended group still yield a valid representation of the dependencies in the Jarret-graph, which at that point can be stated more abstractly, as adhering to the determinant/labels for all instances as in Fig.\ref{fig67}. 

Moreover, $G_{cp}$ has lots of interesting properties and includes several interesting subgroups. To begin with, the first 16 matrices in the group constitute the Pauli group $G_p$, given by 
\beq
G_{p} = \{ \pm {\mathbb I}, \pm X, \pm Y, \pm Z, \pm i {\mathbb I}, \pm iX, \pm iY, \pm iZ \}. 
\eeq
This group is ubiquitous in quantum mechanics, specially in the context of quantum information theory and the study of spin-$1/2$ angular momentum. The group includes matrices $X$, $Y$ and $Z$, which are the generators of the Lie (continuous) group $SU(2)$. This Lie group is the double-cover of the group $SO(3)$ (the group of rotations in $3d$ space), and is also isomorphic to the group of quaternions with norm 1. Moreover, some authors in Ref.\cite{mathphys} have claimed that the Pauli group may play an important role in linguistic theories. It is thus a pleasant surprise that we can actually \emph{derive} this group as a subgroup of $G_{cp}$, starting from a small set of linguistically-motivated axioms. Also, there are many other relevant subgroups in the larger group, including as we saw the ``magnificent eight" cyclic group $G_8 \simeq {\mathbb Z}_2 \times {\mathbb Z}_4$ and all its subgroups, including $G_4 \simeq {\mathbb Z}_4$ and $G_2 \simeq {\mathbb Z}_2$. However, the last 16 matrices in $G_{cp}$ (the ÒChomsky portionÓ) do not form any group by themselves\footnote{  It would be interesting to determine if there is some further structure, e.g., if these matrices are left/right cosets with respect to some subgroup. In this respect, we could not find any obvious relation. \color{black}}.  

$G_{cp}$ needs two generators: for instance $C_1$ and $S_1$. Starting from these two elements, by repeated multiplication one can generate the 32 elements in the group and only them. For completeness, notice that another (perhaps simpler) way of understanding $G_{cp}$ is this: these are the 32 matrices that can be either diagonal or antidiagonal, with $\pm 1$ and $\pm i$ as possible coeffcients. By making the counting, it is easy to see that there can only be 32 such matrices, and these are exactly the ones in $G_{cp}$. Since $\{ \pm 1, \pm i \}$ is also a representation of the cyclic group ${\mathbb Z}_4$, we conclude that the Chomsky-Pauli group $G_{cp}$ is in fact the set of diagonal and antidiagonal $2 \times 2$ matrices with non-zero elements in the group ${\mathbb Z}_4$.

Loosely speaking, the Chomsky-Pauli group $G_{cp}$ is the ``workhorse" of MS as presently explored, in the sense that it suffices, in combination with the other axioms, to build all the relevant structure one needs in order to account for syntactic phenomena in natural language, including long-range linguistic correlations as we have been exploring. 

\subsection{The Chomsky-Pauli Hilbert space $\mathcal{H}_{cp}$} 

Proceeding as with the construction of the Hilbert space $\mathcal{H}_8$ from $G_8$, we now construct the  Hilbert space $\mathcal{H}_{cp}$ associated to the group $G_{cp}$, which we generically call ``Chomsky-Pauli" Hilbert space. One can see that pairs of orthogonal matrices $A$ and $B$ in $G_{cp}$ always obey at least one of the following constraints: 
\begin{enumerate} 
\item{$A$ is diagonal and $B$ antidiagonal, or the other way around (i.e., $A$ antidiagonal and $B$ diagonal).}
\item{$A$ is from the Pauli group and $B$ is also from the Pauli group except $A = \pm B$ and $A = \pm i B$.}
\item{$A$ is not from the Pauli group and $B$ is also not from the Pauli group except $A = \pm B$ and $A = \pm i B$.}
\end{enumerate}Ê
All the scalar products between elements of the group can be derived from Table \ref{tab2}. As was the case for Table \ref{tab1}, rows in the table correspond to the matrix for which there is a $\dagger$ in the scalar product formula of Eq.(\ref{eq3}) (the ``bra"), and columns for those that there is no $\dagger$ (the ``ket"). Products between other elements in $G_{cp}$ are also given by the numbers in the table, up to possible $\pm 1$ and $\pm i$ multiplying factors. 
\begin{table}[h]
\centering
\begin{tabular}{  || c || c | c |c |c |c |c |c |c ||}
\hline
& ${\mathbb I}$  & $X$  & $Y$ & $Z$ & $C_1$ & $C_2$ & $S_1$ & $S_2$ \\ \hline \hline 
${\mathbb I}$ &  $2$ & $0$  & $0$  & $0$ & $1-i$ & $1+i$ & $0$ & $0$ \\\hline
$X$ &  $0$  & $2$ &  $0$           &       $0$          &     $   0 $      &         $  0 $       &     $1-i$  & $1+i$ \\\hline
$Y$ &         $  0$        &         $ 0$         &  $ 2$ &                $  0$          &       $ 0$         &        $ 0$      &      $-1+i$ & $1+i$ \\\hline
$Z$ &        $ 0$      &            $0$          &       $ 0 $       &    $ 2$ &     $1+i$ &   $1-i$ &     $0$      &           $ 0 $         \\\hline
$C_1$ &   $1+i$ &     $ 0 $          &      $ 0 $     &     $ 1-i $&   $2$ &                $ 0 $        &       $  0$      &           $ 0 $         \\\hline 
$C_2$ &    $1-i$ &       $0 $      &         $  0 $      &     $1+i$ &     $ 0 $     &     $  2$ &               $  0$         &         $0$          \\\hline
$S_1$ &          $0$      &      $1+i $&  $-1-i$ &      $  0 $       &         $ 0$       &          $ 0$    &        $ 2$ &             $   0 $         \\\hline
$S_2$ &    $ 0 $    &      $ 1-i$ &  $ 1-i$ &      $  0$        &        $  0$        &        $  0$       &        $   0$           &  $2$  \\\hline
\end{tabular}
\caption{Table of scalar products of some elements in the $G_{cp}$.}
\label{tab2}
\end{table}

By inspecting the above table we see that the set of scalar products of elements in $G_8$ is a subset of those of the elements in $G_{cp}$, which is a natural consequence of $G_8$ being a subgroup of $G_{cp}$. Moreover, we can already see that, according to our definition of scalar product, these tables provide the dimensionality as well as possible choices of basis of a vector space constructed by the elements in the group.

Proceeding as with the derivation of $\mathcal{H}_8$, we now find the dimensionality as well as different basis for $\mathcal{H}_{cp}$. Inspecting Table \ref{tab2} one can see that $\mathcal{H}_{cp}$'s dimension is:
\beq
{\rm dim}(\mathcal{H}_{cp}) = 4. 
\eeq
This is because, according to Table \ref{tab2}, in $G_{cp}$ there are always at most 4 linearly-independent matrices, e.g., the three Pauli matrices together with the identity
\beq
\{  {\mathbb I},  ~ X, ~ Y, ~ Z  \},  
\eeq
but also the four Chomsky matrices
\beq
\{ C_1,  ~ C_2, ~  S_1,  ~ S_2 \}. 
\eeq
To see this, notice that the Chomsky matrices can be written in terms of the three Pauli matrices and the identity as follows: 
\beqa
C_1 = \left( \frac{1-i}{2} \right) {\mathbb I} + \left( \frac{1+i}{2} \right) Z, &~~~& S_1 = \left( \frac{1-i}{2} \right) (X - Y), \nonumber \\Ê
C_2 = \left( \frac{1+i}{2} \right) {\mathbb I} + \left( \frac{1-i}{2} \right) Z, &~~~& S_2 = \left( \frac{1+i}{2} \right) (X + Y) .
\label{change1}
\eeqa
These relations can also be inverted easily, in order to find the three Pauli matrices and the identity from the four Chomsky matrices: 
\beqa
{\mathbb I} = \left( \frac{1+i}{2} \right) C_1 + \left( \frac{1-i}{2} \right) C_2, &~~~& X = \left( \frac{1+i}{2} \right) S_1 + \left( \frac{1-i}{2} \right) S_2 ,\nonumber \\Ê
Z = \left( \frac{1-i}{2} \right) C_1 + \left( \frac{1+i}{2} \right) C_2 ,&~~~&  Y = -\left(\frac{1+i}{2} \right) S_1 + \left( \frac{1-i}{2} \right) S_2.
\label{change2}
\eeqa
As for orthonormal basis, just as was the case for $\mathcal{H}_8$, there are two natural choices for $\mathcal{H}_{cp}$. One is given by the ``normalized" Pauli matrices plus the identity:
\beq
{\rm  Pauli ~ basis} \equiv \left\{  \frac{{\mathbb I}}{\sqrt{2}},  ~ \frac{X}{\sqrt{2}},  ~ \frac{Y}{\sqrt{2}},  ~ \frac{Z}{\sqrt{2}}  \right\} \equiv  \left\{ \ket{\hat{{\mathbb I}}}, ~\ket{\hat{X}}, ~\ket{\hat{Y}}, ~\ket{\hat{Z}} \right\}, 
\eeq
where in the last step we used again the Dirac notation. What this means is that any vector in $\ket{\psi}  \in \mathcal{H}_{cp}$ can be written as 
\beq
\ket{\psi} = \alpha \ket{\hat{I}}  + \beta \ket{\hat{X}}  + \gamma \ket{\hat{Y}}  + \delta \ket{\hat{Z}}, 
\eeq
where $\alpha, \beta, \gamma$ and $\delta$ are complex numbers --- the ``coordinates" of vector $\ket{\psi}$ in this Pauli basis.  The second natural option is given by ``normalized" Chomsky matrices: 
\beq
{\rm  Chomsky ~ basis} \equiv \left\{  \frac{C_1}{\sqrt{2}},  ~ \frac{C_2}{\sqrt{2}},  ~ \frac{S_1}{\sqrt{2}},  ~ \frac{S_2}{\sqrt{2}}  \right\} \equiv  \left\{ \ket{\hat{C_1}}, ~\ket{\hat{C_2}}, ~\ket{\hat{S_1}}, ~\ket{\hat{S_2}} \right\}. 
\eeq
This is an equally valid basis, and any vector in $\mathcal{H}_{cp}$ can be written as 
\beq
\ket{\psi} = \alpha' \ket{\hat{C_1}}  + \beta' \ket{\hat{C_2}}  + \gamma' \ket{\hat{S_1}}  + \delta' \ket{\hat{S_2}}, 
\eeq
where $\alpha', \beta', \gamma'$ and $\delta'$ are the ``coordinates" of vector $\ket{\psi}$ in this Chomsky basis.

Indeed, the relation between these two basis is simply a unitary matrix, that according to Eq.(\ref{change1}) and Eq.(\ref{change2}) can be represented like this: 
\beq
\begin{pmatrix}
  \alpha \\
  \beta \\
 \gamma  \\
  \delta 
 \end{pmatrix} = \begin{pmatrix}
\frac{1-i}{2} & \frac{1+i}{2} & 0 & 0 \\Ê
0 & 0 & \frac{1-i}{2} & \frac{1+i}{2} \\ 
0 & 0 & \frac{i-1}{2} & \frac{1+i}{2} \\ 
\frac{1+i}{2} & \frac{1-i}{2} & 0 & 0 
\end{pmatrix} \begin{pmatrix}
  \alpha' \\
  \beta' \\
 \gamma'  \\
  \delta' 
 \end{pmatrix}.
 \eeq
The above matrix-vector multiplication gives the rule to change from ``Chomsky coordinates" to ``Pauli coordinates" for any vector in the Hilbert space $\mathcal{H}_{cp}$. Both ``coordinate systems" are valid, as far as the description of the space is concerned. Other basis of the Hilbert space  $\mathcal{H}_{cp}$ are of course possible. 

Finally, notice that the Hilbert space $\mathcal{H}_8$ is a $2d$ subspace of $\mathcal{H}_{cp}$, in the same way that $G_8$ is a subgroup of $G_{cp}$. In fact, we have that 
\beq
\mathcal{H}_{cp} \simeq {\mathbb C}^2 \oplus  {\mathbb C}^2,  
\eeq
i.e., the more general $4d$ space $\mathcal{H}_{cp}$ is in fact isomorphic to the Hilbert space of the direct sum of two qubits. One of the qubits is described by space $\mathcal{H}_8$, spanned by vectors $\ket{\hat{{\mathbb I}}}$ and $\ket{\hat{Z}}$ (or equivalently by $\ket{\hat{C_1}}$ and $\ket{\hat{C_2}}$), and the other corresponds to the $2d$ subspace spanned by the vectors $\ket{\hat{X}}$ and $\ket{\hat{Y}}$ (or equivalently by $\ket{\hat{S_1}}$ and $\ket{\hat{S_2}}$). As such, $G_{cp}$ and its associated Hilbert space $\mathcal{H}_{cp}$ provide much more room to account for syntax derivations than $G_8$ and $\mathcal{H}_8$. It is intriguing that this structure entails also the Pauli matrices as a natural consequence. 

The group $G_{cp}$ and its associated Hilbert space $\mathcal{H}_{cp}$ are the biggest mathematical structures so far in our theory. Most of the derivations could be carried out with $G_8$ and $\mathcal{H}_8$ only but, as said above, more structure is indeed needed in realistic settings. As an example, in the next two sections we show how the structure of $\mathcal{H}_{cp}$ indeed allows for non-trivial chain conditions, not all of them showing up if one considers $\mathcal{H}_8$ only, and which need to be analyzed in detail.

\clearpage 

\part{Chain conditions in more detail}
\label{p3}

\section{Unit matrices and their role} 
\label{sec12b}
In Secs.\ref{sec10}Ê and \ref{sec11}Ê we have seen the basics of how linguistic chains show up in our MS model. The purpose of this section is to delve into the details, putting at work all the mathematical machinery that we have presented. For the sake of generality, the derivations in this section are done using the group $G_{cp}$,  and we deal with the group $G_{8}$ as a particular case of the general setting. \color{black}Ê

First of all, let us come back to this question: in our Hilbert space $\mathcal{H}_{cp}$, which chain constructions are valid? For the purpose of this section, we demand chains to follow these natural and practical constraints: 
\begin{enumerate}
\item{They must be compressible\footnote{ÊIn the sense that they must have some null eigenspace. \color{black}} back to $2 \times 2$ matrices.}
\item{The resulting $2 \times 2$ matrix must have a label (i.e., complex phase of the determinant) equal to $\pm 1, \pm i$.}
\item{The chain is formed by summing perpendicular vectors in $\mathcal{H}_{cp}$ (i.e., superposition of orthogonal contexts).}Ê
\end{enumerate}
The first condition is a restriction on what is imposed by Axiom 11. The second one follows  from Axiom  7 and the Projection Theorem. The third one is an \emph{a priori} reasonable working assumption.  Notice that this third condition does not follow from the axioms of MS themselves, and is an extra assumption that we may choose to work with or not. 

As explained in Sec.\ref{sec10}, a chain is given by Eq.(\ref{chain1}) in Dirac notation. Let us refresh it: 
\beq
\ket{{\rm Chain}} = \frac{1}{\sqrt{2}} \left(\ket{\hat{\alpha}} \otimes  \ket{\hat{K}}
 + \ket{\hat{\alpha}} \otimes  \ket{\hat{\Lambda}} \right) =  \frac{1}{\sqrt{2}} \ket{\hat{\alpha}} \otimes \left( \ket{\hat{K}} + \ket{\hat{\Lambda}} \right). 
\eeq
In this equation, $\alpha, K$ and $\Lambda$ are matrices from the group $G_{cp}$, where $\alpha$ denotes the specifier and $K, \Lambda$ the relevant ``movement contexts". Our first requirement implies that the matrix sum $K + \Lambda$ \emph{necessarily} has one and only one zero eigenvalue. This needs to be so, because matrix $\alpha \in G_{cp}$ always has two nonzero eigenvalues, and the eigenvalues of $\alpha \otimes \left( K + \Lambda \right)$ are the 4 product combinations of the two eigenvalues coming from $\alpha$ and the two coming from $ K + \Lambda $. Therefore, if the matrix resulting from the chain must have 2 non-zero eigenvalues, then this can only happen if one of the eigenvalues of $K + \Lambda$ is null. 

By looking at all possible cases, it is easy to see that there are exactly 96 combinations $U = A + B$, with $A, B \in G_{cp}$, $A \neq B$, and such that $U$ has one non-zero eigenvalue only. We call these 96 combinations the \emph{unit matrices}\footnote{Not to be confused with unitary matrices, which are defined as those matrices $U$ satisfying $U U^{\dagger} = U^{\dagger} U = \mathbb{I}$.}. In terms of such matrices, a chain reads as follows: 
\beq
{\rm Chain} =  \alpha \otimes A + \alpha \otimes B = \alpha \otimes \left( A + B \right) = \alpha \otimes U. 
\eeq
The above matrix is compressible back to a $2 \times 2$ matrix, since $\alpha$ is $2 \times 2$ and $U$ is compressible to a $1 \times 1$ diagonal matrix (which implies that it is proportional to a projector). Let us call these compressed $1 \times 1$ matrices $U_c$ (accounting for ``compressed $U$"). The label of the chain formed by the $U$ matrices as context, after compression (i.e. with $U_c$), follows the usual rules of the determinant of a tensor product and, importantly, a \emph{rescaling} by the norm. This is given by: 
\beq
{\rm Label(Chain)} = \frac{1}{N} \det \left( \alpha \otimes U_c \right) = \frac{1}{N}\left( \det\left( \alpha \right) \right) \left( \det\left(U_c \right) \right)^2,   
\eeq
with the rescaling factor
\beq
N \equiv \left|\left( \det\left( \alpha \right) \right) \left( \det\left(U_c \right) \right)^2 \right| =  \left|\det\left(U_c \right)\right|^2, 
\label{rescaling}
\eeqÊ
i.e., the \emph{phase in the complex plane} of $\left( \det\left( \alpha \right) \right) \left( \det\left(U_c \right) \right)^2$ is the label of the chain, and the \emph{norm} of this same quantity is the rescaling factor. The last equality in Eq.(\ref{rescaling}) follows from the fact that $\det \left( \alpha \right)$ is always equal to $\pm1, \pm i$ if $\alpha \in G_{cp}$, and has therefore norm $1$. 

Two remarks are in order: first, \emph{the chain label is the specifier label times the square of the $U_c$ label, with a rescaling by its norm}. Second, we will see in what follows that the rescaling is indeed necessary in order to keep consistency of the obtained linguistic labels. The introduction of this rescaling may look like a bit artificial to a linguist. However, at a physical level it makes perfect sense: it amounts to \emph{the rescaling of linguistic variables after a coarse-graining of linguistic information, thus constituting a full renormalization step}. It is remarkable that, without looking for it, we just found the interpretation discussed in Ref.\cite{gallegoorus}Ê of the operation M as a coarse-graining step, with the subsequent link to renormalization in physics. As physicists know well, a renormalization step is always built by a coarse-graining followed by a rescaling of the physical variables. Such a rescaling is needed in order to have the renormalized object defined in the same units than the original unrenormalized one. So far we knew about the coarse-graining (which in Ref.\cite{gallegoorus} is conjectured to be M in linguistics), but now we just found a non-trivial example of the second ingredient. The rescaling of the physical variables is, precisely, the rescaling needed in Eq.(\ref{rescaling}) in order to recover the original set of labels $\pm 1, \pm i$, which makes our theory mathematically consistent. 

Let us now take a closer look at the 96 combinations producing the $U$-matrices. It turns out that these can be divided into 3 sets of 32 combinations each, which we give in Tables \ref{Set1}, \ref{Set2} and \ref{Set3}. ÊThe 16 combinations arising from the group $G_8$ are a particular case and are highlighted in the tables with a smiley  {\bf \smiley}). \color{black}Ê
 
 %%%%%

 \begin{table}
\centering
\begin{tabular}{ || c ||  c | c | c | c ||}
\hline
  & \multicolumn{4}{c||}{} \\ 
$(\det(U_c))^2/N$ & \multicolumn{4}{c||}{$U$ matrices}\\ 
 & \multicolumn{4}{c||}{} \\ \hline
\hline
  & &  &  &   \\ 
& $\mathbb{I} +X$  & $\mathbb{I} + Y$ & {\bf \smiley} $\mathbb{I} + Z$ & $\mathbb{I}-X$  \\ 
&  &  &  &   \\ 
$1$& 
$\begin{pmatrix}
1 & 1 \\Ê
1 & 1  
\end{pmatrix}$ 
& 
$\begin{pmatrix}
1 & -i \\Ê
i & 1  
\end{pmatrix}$ 
& 
$\begin{pmatrix}
2 & 0 \\Ê
0 & 0  
\end{pmatrix}$ 
& 
$\begin{pmatrix}
1 & -1 \\Ê
-1 & 1  
\end{pmatrix}$ \\
&  &  &  &   \\\hline
   & &  &  &   \\ 
    & $ \mathbb{I} - Y$  & {\bf \smiley} $\mathbb{I} - Z$ & {\bf \smiley} $C_1+C_2$ & $iC_1-iC_2$  \\ 
 & &  &  &   \\ 
 $1$&
$\begin{pmatrix}
1 & i \\Ê
-i & 1  
\end{pmatrix}$ 
& 
$\begin{pmatrix}
0 & 0 \\Ê
0 & 2  
\end{pmatrix}$ 
& 
$\begin{pmatrix}
2 & 0 \\Ê
0 & 0  
\end{pmatrix}$ 
& 
$\begin{pmatrix}
0 & 0 \\Ê
0 & 2  
\end{pmatrix}$ \\
&  &  &  &   \\\hline
   & &  &  &   \\ 
& $-\mathbb{I} + X$  & $-\mathbb{I} +Y$ & {\bf \smiley}Ê $-\mathbb{I}+Z$ & $-\mathbb{I}-X$  \\ 
&  &  &  &   \\ 
$1$ &
$\begin{pmatrix}
-1 & 1 \\Ê
1 & -1  
\end{pmatrix}$ 
& 
$\begin{pmatrix}
-1 & -i \\Ê
i & -1  
\end{pmatrix}$ 
& 
$\begin{pmatrix}
0 & 0 \\Ê
0 & -2  
\end{pmatrix}$ 
& 
$\begin{pmatrix}
-1 & -1 \\Ê
-1 & -1  
\end{pmatrix}$ \\
 & &  &  &   \\\hline
 &   &  &  &   \\ 
 & $-\mathbb{I}-Y$  & {\bf \smiley} $-\mathbb{I}-Z$ & {\bf \smiley} $-C_1-C_2$ & $-iC_1+iC_2$  \\ 
&  &  &  &   \\ 
$1$&
$\begin{pmatrix}
-1 & i \\Ê
-i & -1  
\end{pmatrix}$ 
& 
$\begin{pmatrix}
-2 & 0 \\Ê
0 & 0  
\end{pmatrix}$ 
& 
$\begin{pmatrix}
-2 & 0 \\Ê
0 & 0  
\end{pmatrix}$ 
& 
$\begin{pmatrix}
0 & 0 \\Ê
0 & -2  
\end{pmatrix}$ \\
&  &  &  &   \\\hline
  & &  &  &   \\ 
 & $i \mathbb{I} + iX$  & $i\mathbb{I} +iY$ & $i\mathbb{I}+iZ$ & $i\mathbb{I}-iX$  \\ 
&  &  &  &   \\ 
$-1$&
$\begin{pmatrix}
i & i \\Ê
i & i  
\end{pmatrix}$ 
& 
$\begin{pmatrix}
i & 1 \\Ê
-1 & i  
\end{pmatrix}$ 
& 
$\begin{pmatrix}
2i & 0 \\Ê
0 & 0  
\end{pmatrix}$ 
& 
$\begin{pmatrix}
i & -i \\Ê
-i & i  
\end{pmatrix}$ \\
&  &  &  &   \\\hline
 &   &  &  &   \\ 
 & $i\mathbb{I} - iY$  & $i\mathbb{I} - iZ$ & {\bf \smiley} $-C_1+C_2$ & $iC_1+iC_2$  \\ 
&  &  &  &   \\ 
$-1$&
$\begin{pmatrix}
i & -1 \\Ê
1 & i  
\end{pmatrix}$ 
& 
$\begin{pmatrix}
0 & 0 \\Ê
0 & 2i  
\end{pmatrix}$ 
& 
$\begin{pmatrix}
0 & 0 \\Ê
0 & 2i  
\end{pmatrix}$ 
& 
$\begin{pmatrix}
2i & 0 \\Ê
0 & 0  
\end{pmatrix}$ \\
&  &  &  &   \\\hline
  &  &  &  &   \\ 
 & $-i\mathbb{I} +iX$  & $-i\mathbb{I} + iY$ & $-i\mathbb{I}+iZ$ & $-i\mathbb{I}-iX$  \\ 
&  &  &  &   \\ 
$-1$&
$\begin{pmatrix}
-i& i \\Ê
i & -i  
\end{pmatrix}$ 
& 
$\begin{pmatrix}
-i & 1 \\Ê
-1 & -i  
\end{pmatrix}$ 
& 
$\begin{pmatrix}
0 & 0 \\Ê
0 & -2i  
\end{pmatrix}$ 
& 
$\begin{pmatrix}
-i & -i \\Ê
-i & -i  
\end{pmatrix}$ \\
&  &  &  &   \\\hline
  &  &  &  &   \\ 
 & $-i\mathbb{I} - iY$  & $-i\mathbb{I}-iZ$ & {\bf \smiley} $C_1-C_2$ & $-iC_1-iC_2$  \\ 
&  &  &  &   \\ 
$-1$&
$\begin{pmatrix}
-i & -1 \\Ê
1 & -i  
\end{pmatrix}$ 
& 
$\begin{pmatrix}
-2i & 0 \\Ê
0 & 0  
\end{pmatrix}$ 
& 
$\begin{pmatrix}
0 & 0 \\Ê
0 & -2i  
\end{pmatrix}$ 
& 
$\begin{pmatrix}
-2i & 0 \\Ê
0 & 0  
\end{pmatrix}$ \\
&   &  &  &   \\\hline  
\end{tabular}
\caption{First set of unit matrices. The ones obtained from $G_8$ are marked with a smiley ({\bf \smiley}). }
\label{Set1}
\end{table}

%%%%

 \begin{table}
\centering
\begin{tabular}{ || c ||  c | c | c | c ||}
\hline
  & \multicolumn{4}{c||}{} \\ 
$(\det(U_c))^2/N$ & \multicolumn{4}{c||}{$U$ matrices}\\ 
 & \multicolumn{4}{c||}{} \\ \hline
\hline
  & &  &  &   \\ 
   & $S_2 + C_2$  & $-S_2 + C_2$ & $S_2-C_2$ & $-S_2-C_2$  \\ 
&  &  &  &   \\ 
$i$&
$\begin{pmatrix}
1 & 1 \\Ê
i & i  
\end{pmatrix}$ 
& 
$\begin{pmatrix}
1 & -1 \\Ê
-i & i  
\end{pmatrix}$ 
& 
$\begin{pmatrix}
-1 & 1 \\Ê
i & -i  
\end{pmatrix}$ 
& 
$\begin{pmatrix}
-1 & -1 \\Ê
-i & -i  
\end{pmatrix}$ \\
&  &  &  &   \\\hline
 &   &  &  &   \\ 
 & $iS_1 + C_2$  & $-iS_1 - C_2$ & $iS_1-C_2$ & $-iS_1+C_2$  \\ 
&  &  &  &   \\ 
$i$&
$\begin{pmatrix}
1 & i \\Ê
1 & i  
\end{pmatrix}$ 
& 
$\begin{pmatrix}
-1 & -i \\Ê
-1 & -i  
\end{pmatrix}$ 
& 
$\begin{pmatrix}
-1 & i \\Ê
1 & -i  
\end{pmatrix}$ 
& 
$\begin{pmatrix}
1 & -i \\Ê
-1 & i  
\end{pmatrix}$ \\
&  &  &  &   \\\hline
  &  &  &  &   \\ 
 & $iS_1 + iC_1$  & $-iS_1 - iC_1$ & $-iS_1+iC_1$ & $iS_1-iC_1$  \\ 
&  &  &  &   \\ 
$i$&
$\begin{pmatrix}
i & i \\Ê
1 & 1  
\end{pmatrix}$ 
& 
$\begin{pmatrix}
-i & -i \\Ê
-1 & -1  
\end{pmatrix}$ 
& 
$\begin{pmatrix}
i & -i \\Ê
-1 & 1  
\end{pmatrix}$ 
& 
$\begin{pmatrix}
-i & i \\Ê
1 & -1  
\end{pmatrix}$ \\
&  &  &  &   \\\hline
  &  &  &  &   \\ 
 & $S_2 + iC_1$  & $-S_2 - iC_2$ & $S_2-iC_1$ & $-S_2-iC_1$  \\ 
&  &  &  &   \\ 
$i$&
$\begin{pmatrix}
i & 1 \\Ê
i & 1  
\end{pmatrix}$ 
& 
$\begin{pmatrix}
-i & -1 \\Ê
-i & 1  
\end{pmatrix}$ 
& 
$\begin{pmatrix}
-i & 1 \\Ê
i & -1  
\end{pmatrix}$ 
& 
$\begin{pmatrix}
-i & -1 \\Ê
-i & -1  
\end{pmatrix}$ \\
&   &  &  &   \\\hline  
  &  &  &  &   \\ 
& $S_1 + C_1$  & $-S_1 - C_1$ & $S_1-C_1$ & $-S_1+C_1$  \\ 
&  &  &  &   \\ 
$-i$& 
$\begin{pmatrix}
1 & 1 \\Ê
-i & -i  
\end{pmatrix}$ 
& 
$\begin{pmatrix}
-1 & -1 \\Ê
i & i  
\end{pmatrix}$ 
& 
$\begin{pmatrix}
-1 & 1 \\Ê
-i & i  
\end{pmatrix}$ 
& 
$\begin{pmatrix}
1 & -1 \\Ê
i & -i  
\end{pmatrix}$ \\
&  &  &  &   \\\hline
   & &  &  &   \\ 
    & $iS_2 + C_1$  & $iS_2 - C_1$ & $-iS_2+C_1$ & $-iS_2-C_1$  \\ 
 & &  &  &   \\ 
 $-i$&
$\begin{pmatrix}
1 & i \\Ê
-1 & -i  
\end{pmatrix}$ 
& 
$\begin{pmatrix}
-1 & i \\Ê
-1 & i  
\end{pmatrix}$ 
& 
$\begin{pmatrix}
1 & -i \\Ê
1 & -i  
\end{pmatrix}$ 
& 
$\begin{pmatrix}
-1 & -i \\Ê
1 & i  
\end{pmatrix}$ \\
&  &  &  &   \\\hline
   & &  &  &   \\ 
& $iS_2 + iC_2$  & $-iS_2 - iC_2$ & $-iS_2+iC_2$ & $iS_2-iC_2$  \\ 
&  &  &  &   \\ 
$-i$ &
$\begin{pmatrix}
i & i \\Ê
-1 & -1  
\end{pmatrix}$ 
& 
$\begin{pmatrix}
-i & -i \\Ê
1 & 1  
\end{pmatrix}$ 
& 
$\begin{pmatrix}
i & -i \\Ê
1 & -1  
\end{pmatrix}$ 
& 
$\begin{pmatrix}
-i & i \\Ê
-1 & 1  
\end{pmatrix}$ \\
 & &  &  &   \\\hline
 &   &  &  &   \\ 
 & $S_1 + iC_2$  & $-S_1 - iC_2$ & $S_1-iC_2$ & $-S_1+iC_2$  \\ 
&  &  &  &   \\ 
$-i$&
$\begin{pmatrix}
i & 1 \\Ê
-i & -1  
\end{pmatrix}$ 
& 
$\begin{pmatrix}
-i & -1 \\Ê
i & 1  
\end{pmatrix}$ 
& 
$\begin{pmatrix}
-i & 1 \\Ê
-i & 1  
\end{pmatrix}$ 
& 
$\begin{pmatrix}
i & -1 \\Ê
i & -1  
\end{pmatrix}$ \\
&  &  &  &   \\\hline
\end{tabular}
\caption{Second set of unit matrices.}
\label{Set2} 
\end{table}

%%%%%

 \begin{table}
\centering
\begin{tabular}{ || c ||  c | c | c | c ||}
\hline
  & \multicolumn{4}{c||}{} \\ 
$(\det(U_c))^2/N$ & \multicolumn{4}{c||}{$U$ matrices}\\ 
 & \multicolumn{4}{c||}{} \\ \hline
\hline
  & &  &  &   \\ 
& {\bf \smiley} $\mathbb{I} - C_1$  & {\bf \smiley} $-\mathbb{I} + C_1$ & $\mathbb{I} + iC_2$ & $-\mathbb{I}+iC_2$  \\ 
&  &  &  &   \\ 
$i$& 
$\begin{pmatrix}
0 & 0 \\Ê
0 & 1+i  
\end{pmatrix}$ 
& 
$\begin{pmatrix}
0 & 0 \\Ê
0 & -1-i  
\end{pmatrix}$ 
& 
$\begin{pmatrix}
1+i & 0 \\Ê
0 & 0  
\end{pmatrix}$ 
& 
$\begin{pmatrix}
-1-i & 0 \\Ê
0 & 0  
\end{pmatrix}$ \\
&  &  &  &   \\\hline
  & &  &  &   \\ 
 & $i \mathbb{I} + C_1$  & $-i\mathbb{I} - C_1$ & $i\mathbb{I}-iC_2$ & $-i\mathbb{I}+iC_2$  \\ 
&  &  &  &   \\ 
$i$&
$\begin{pmatrix}
1+i & 0 \\Ê
0 & 0  
\end{pmatrix}$ 
& 
$\begin{pmatrix}
-1-i & 0 \\Ê
0 & 0  
\end{pmatrix}$ 
& 
$\begin{pmatrix}
0 & 0 \\Ê
0 & 1+i  
\end{pmatrix}$ 
& 
$\begin{pmatrix}
0 & 0 \\Ê
0 & -1-i  
\end{pmatrix}$ \\
&  &  &  &   \\\hline
 & &  &  &   \\ 
    & {\bf \smiley} $ Z - C_2$  & {\bf \smiley} $-Z + C_2$ & $-Z-iC_1$ & $Z+iC_1$  \\ 
 & &  &  &   \\ 
 $i$&
$\begin{pmatrix}
0 & 0 \\Ê
0 & -1-i  
\end{pmatrix}$ 
& 
$\begin{pmatrix}
0 & 0 \\Ê
0 & 1+i  
\end{pmatrix}$ 
& 
$\begin{pmatrix}
-1-i & 0 \\Ê
0 & 0  
\end{pmatrix}$ 
& 
$\begin{pmatrix}
1+i & 0 \\Ê
0 & 0  
\end{pmatrix}$ \\
&  &  &  &   \\\hline
 &   &  &  &   \\ 
 & $iZ + C_2$  & $-iZ - C_2$ & $iZ-iC_1$ & $-iZ+iC_1$  \\ 
&  &  &  &   \\ 
$i$&
$\begin{pmatrix}
1+i & 0 \\Ê
0 & 0  
\end{pmatrix}$ 
& 
$\begin{pmatrix}
-1-i & 0 \\Ê
0 & 0  
\end{pmatrix}$ 
& 
$\begin{pmatrix}
0 & 0 \\Ê
0 & -1-i  
\end{pmatrix}$ 
& 
$\begin{pmatrix}
0 & 0 \\Ê
0 & 1+i  
\end{pmatrix}$ \\
&  &  &  &   \\\hline
& &  &  &   \\ 
& {\bf \smiley} $\mathbb{I} - C_2$  & {\bf \smiley} $-\mathbb{I} + C_2$ & $\mathbb{I}-iC_1$ & $-\mathbb{I}+iC_1$  \\ 
&  &  &  &   \\ 
$-i$ &
$\begin{pmatrix}
0 & 0 \\Ê
0 & 1-i  
\end{pmatrix}$ 
& 
$\begin{pmatrix}
0 & 0 \\Ê
0 & -1+i  
\end{pmatrix}$ 
& 
$\begin{pmatrix}
1-i & 0 \\Ê
0 & 0  
\end{pmatrix}$ 
& 
$\begin{pmatrix}
-1+i & 0 \\Ê
0 & 0  
\end{pmatrix}$ \\
 & &  &  &   \\\hline
 &  &  &  &   \\ 
 & $i\mathbb{I} - C_2$  & $-i\mathbb{I} + C_2$ & $i\mathbb{I}-iC_1$ & $-i\mathbb{I}+iC_1$  \\ 
&  &  &  &   \\ 
$-i$&
$\begin{pmatrix}
-1+i& 0 \\Ê
0 & 0  
\end{pmatrix}$ 
& 
$\begin{pmatrix}
1-i & 0 \\Ê
0 & 0  
\end{pmatrix}$ 
& 
$\begin{pmatrix}
0 & 0 \\Ê
0 & -1+i  
\end{pmatrix}$ 
& 
$\begin{pmatrix}
0 & 0 \\Ê
0 & 1-i  
\end{pmatrix}$ \\
&  &  &  &   \\\hline
 &   &  &  &   \\ 
 & {\bf \smiley} $Z -C_1$  & {\bf \smiley} $-Z + C_1$ & $-Z+iC_2$ & $Z-iC_2$  \\ 
&  &  &  &   \\ 
$-i$&
$\begin{pmatrix}
0 & 0 \\Ê
0 & -1+i  
\end{pmatrix}$ 
& 
$\begin{pmatrix}
0 & 0 \\Ê
0 & 1-i  
\end{pmatrix}$ 
& 
$\begin{pmatrix}
-1+i & 0 \\Ê
0 & 0  
\end{pmatrix}$ 
& 
$\begin{pmatrix}
1-i & 0 \\Ê
0 & 0  
\end{pmatrix}$ \\
&  &  &  &   \\\hline
  &  &  &  &   \\ 
 & $iZ - C_1$  & $-iZ+C_1$ & $iZ-iC_2$ & $-iZ+iC_2$  \\ 
&  &  &  &   \\ 
$-i$&
$\begin{pmatrix}
-1+i & 0 \\Ê
0 & 0  
\end{pmatrix}$ 
& 
$\begin{pmatrix}
1-i & 0 \\Ê
0 & 0  
\end{pmatrix}$ 
& 
$\begin{pmatrix}
0 & 0 \\Ê
0 & 1-i  
\end{pmatrix}$ 
& 
$\begin{pmatrix}
0 & 0 \\Ê
0 & -1+i  
\end{pmatrix}$ \\
&   &  &  &   \\\hline  
\end{tabular}
\caption{Third set of unit matrices. The ones obtained from $G_8$ are marked with a smiley ({\bf \smiley}).}
\label{Set3}
\end{table}

We now consider the defining properties of these three sets. In Set 1 (Table \ref{Set1}), all matrices correspond to sums of orthogonal contexts, and moreover those with ``context label"  $(\det(U_c))^2/N = +1$ are \emph{hermitian}, and those with $- 1$ are \emph{anti-hermitian}. Additionally, in this set there are exactly $8$ matrices each with a unique non-zero eigenvalue proportional to $\pm 1, \pm i$. In Set 2 (Table \ref{Set2}), all matrices correspond to sums of \emph{orthogonal} contexts. In this set there are also exactly $8$ matrices each with a unique non-zero eigenvalue proportional to $(\pm 1 \pm i)$ --- i.e., $1 + i, 1 - i, -1 + i, -1 - i$. In Set 3 (Table \ref{Set3}), all matrices correspond to sums of \emph{non-orthogonal} contexts, all are \emph{unitary} Ê(which as noted on footnote 17 has nothing to do with matrices being units in our terms) \color{black}Êand, additionally, they are diagonal. Again, in this set there are exactly $8$ matrices each with a unique non-zero eigenvalue proportional to $(\pm 1 \pm i)$. 

As per the constraints mentioned at the beginning of this section, Set 3 does not fulfil the orthogonality condition. We are therefore left with Sets 1 and 2 to describe reasonable chains. This is particularly appealing, since then there are exactly 16 possible combinations leading to each one of the 4 possible values $\pm 1, \pm i$ for the ``context label" . Notice, though, that as per our formalism, all the matrices in these three sets are in principle possible in the construction of a chain. We leave it for future work, probably based on linguistic constraints, to determine if all these combinations are indeed relevant to describe reasonable chains, or if only some subset of them is the one that makes linguistic sense. \emph{A priori} there is nothing against Set 3 apart from the orthogonality constraint. However, we find that the symmetries present in Sets 1 and 2 are interesting: first, all lexical labels show up there equally distributed in the ``context label", and second, Set 1 is formed by hermitian and anti-hermitian matrices, which we find \emph{aesthetically and logically}  appealing. 

In a nutshell: 

\bigskip
$\bullet$ \emph{{\bf Conjecture:} Êthe relevant context superpositions in reasonable linguistic chains are the ones in Sets 1 and 2.}
\bigskip

What about the possible group structure of chain matrices? Well, without compression, it is clear that matrices $U$ \emph{never} form a group by themselves, simply because the $2 \times 2$ identity matrix is not in the set as it has two non-zero eigenvalues instead of only one (as required to be a $U$ matrix). After compression, however, the story is different. With a careful mathematical analysis one can see that, overall, the resulting matrices $\alpha \otimes U_c$ do not form a group in general, since the projectors for the different $U_c$ are not always orthogonal (e.g., consider those for $\mathbb{I} + X$ and $\mathbb{I} + Z$). However, it is easy to identify \emph{subsets} of unit matrices which, if considered individually and separate from the rest of $U$ matrices, then they do have a group structure. This is the case, for instance, of the subset $\{ \mathbb{I} + X, -\mathbb{I} - X, i \mathbb{I} + iX, -i \mathbb{I} -i X \}$: those four matrices have a unique non-zero eigenvalue equal to $\pm 1, \pm i$, and the four of them have the same corresponding eigenvector.  As a matter of fact, one can see that \emph{all} the matrices in the three sets belong to some subset of 4 matrices obeying similar conditions, though sometimes with context labels $1 + i, 1 - i, -1 + i, -1 - i$  Ê(As remarked above, the key is to identify sets of matrices with the same eigenvector). \color{black} If only $U$ matrices within one of these subsets are considered, then the corresponding compressed (restricted) chains \emph{do have} a group structure, which is governed by the group 
\beq
G_{{\rm r-chain}} = G_{cp} \otimes \mathbb{Z}_4. 
\eeq
In the equation above, by ``r-chain" we mean ``restricted chain", i.e., a chain formed with the 4 $U$-matrices in one of the subsets mentioned above.  The first term in $G_{{\rm r-chain}}$ is $G_{cp}$, i.e., the group of matrices $\alpha$. The second term is just the considered set of 4 $U_c$ matrices, proportional to $1 \times 1$ projectors, which constitute nothing but a representation of the group $\mathbb{Z}_4$, i.e., the group of the integers modulo 4. The matrices in the group $G_{{\rm r-chain}}$ are therefore all the possible \emph{tensor products} of the matrices in  $G_{cp}$ and the considered representation of $\mathbb{Z}_4$. Such ``r-chains" constitute our best candidates for the workable chains in language. 

Let us conclude this section by saying that a more detailed analysis of chain conditions in MS from purely linguistic grounds, with particular examples in real-world sentences, will appear in future works for audiences with a formal linguistic background. We wish, however, to discuss a least one explicit example in the following section. 

\section{An actual example with presuppositions and entailments}Ê
\label{newsec}
Consider again an arbitrary chain state as in Eq.(\ref{archain}):
\beq
\alpha \otimes K + \alpha \otimes \Lambda = \alpha \left( K + \Lambda \right), ~~~~ {\rm with~} \alpha, K, \Lambda \in G_{pc}. 
\label{archain}
\eeq
Let $\alpha$ be a specifier and $K$, $\Lambda$ relevant ``movement contexts". Such specification operations lead to computational blowup, so in the previous section we restricted ourselves to 64 $2 \times 2$ ``unit" ($U$) matrix combinations that allow compression. Broadly, these comprise all possible ``movement contexts" for the specifiers of the chains we are studying. These are ``possible" in a grammatical sense, including scruples about the chain's label. To meet those, we are assuming chains whose label corresponds to all four primitive labels, what we have called r-chains. 

In the previous section we have seen three groups of $U$ matrices, as presented in Tables \ref{Set1}, \ref{Set2} and \ref{Set3}, three rather different sets of 32 elements each. Remembering that \emph{the chain label is the specifier label times the square of the $U$ label (normalized)}, observe how the 32 $U$ matrices in Table \ref{Set1} divide into two sets. 16 of those have the property $( \det( U_c))^2/N = 1$ and 16 more, in contrast, the property  $( \det( U_c))^2/N = -1$. These are the only 32 matrices within the entire set of 96 $U$ matrices with label $\pm 1$ --- the ones with label 1 being hermitian and the ones with label -1 anti-hermitian. For these 32, \emph{the specifier label is identical to the putative chain label or its polar opposite}. We may call these \emph{harmonic} combinations. The other 64 matrices are non-harmonic in the sense just described. 32 of those (combined) 64 matrices, namely half of those in Tables \ref{Set2} and \ref{Set3}, have the property $( \det( U_c))^2/N = i$ and the other 32, the property  $( \det( U_c))^2/N = -i$. For these less elegant 64 matrices, \emph{the specifier label is in no simple relation to the putative chain label}.

Just as the Jarret graph tells us which first mergers work (given ``projection scruples"), we need to sieve through the possible labeled chains to see which deliver: a) compressed specifiers that preserve their (pre-compression) label, and b) chains whose label is appropriate for the selection conditions (if any) of higher material relating to such chains. In that respect, the observation just made about the correlation between a specifier and a chain label (harmonic vs. non-harmonic combinations) is important. In general, standard symmetrical properties ought to guide structural possibilities. Note how the (anti-)hermitian $U$ matrices in Table \ref{Set1} involve \emph{orthogonal} contexts, a condition \emph{sine qua non} for the formation of chains via the superposition of chain occurrences. It is natural to try and map the most elegant elements to standard chains. We call the (anti-)hermitian matrices the ``thirty two marks" of long-range dependency.

As we try to identify the grammatical role that these ``thirty two marks" of grammatical chain contexts may be playing, it is useful to separate them into the following three sets: 

\begin{enumerate}
\item{\emph{Predicational Contexts:}
\begin{table}[h]
\centering
\begin{tabular}{c c c c c c c}
$~~~{\rm det } = 1$: &~ $\mathbb{I}Ê+ X$, &~~$\mathbb{I} + Y$, &~~$\mathbb{I} + Z$, & ~~$\mathbb{I}Ê- X$, &~~$\mathbb{I} - Y$, &~$\mathbb{I} - Z$  \\Ê
${\rm det } = -1$: &$-\mathbb{I}Ê+ X$, &$-\mathbb{I} + Y$, &$-\mathbb{I} + Z$, & $-\mathbb{I}Ê- X$, &$-\mathbb{I} - Y$, &$-\mathbb{I} - Z$ Ê
\end{tabular}Ê
\end{table}}
\item{\emph{Movement Contexts:}
\begin{table}[h]
\centering
\begin{tabular}{c c c c c c c}
$~~~{\rm det } = i$: &~ $i\mathbb{I}Ê+ iX$, &~~$i\mathbb{I} + iY$, &~~$i\mathbb{I} + iZ$, & ~~$i\mathbb{I}Ê- iX$, &~~$i\mathbb{I} - iY$, &~$i\mathbb{I} - iZ$  \\Ê
${\rm det } = -i$: &$-i\mathbb{I}Ê+ iX$, &$-i\mathbb{I} + iY$, &$-i\mathbb{I} + iZ$, & $-i\mathbb{I}Ê- iX$, &$-i\mathbb{I} - iY$, &$-i\mathbb{I} - iZ$ Ê
\end{tabular}Ê
\end{table}}
\item{\emph{Head Contexts:}
\begin{table}[h]
\centering
\begin{tabular}{c c }
$~[ C_1 + C_2]_A$, & $[iC_1 - iC_2]_1$ \\Ê
$~~[ -C_1 - C_2]_N$, & $~~~~[-iC_1 + iC_2]_{-1}$ \\Ê
$~~[ -C_1 + C_2]_V$, & $[iC_1 + iC_2]_i$ \\Ê
$[ C_1 - C_2]_P$, & $~~~~[-iC_1 - iC_2]_{-i}$ 
\end{tabular}Ê
\end{table}}
\end{enumerate} 
Take the set in (3). These combinations make no sense at a phrasal level: depending on whether we combine either one of the ``twin" varieties (matrices with the exact same determinant and diagonal format, except with opposite values in non-zero terms), we get totally different determinants, and this is not what we expect out of phrasal combinations. But as head combinations we can explore what they may mean. In contrast, the \emph{Predicational} (1) and \emph{Movement} (2) sets do make sense in that both ``twin" categories act consistently in terms of their determinant-label.  

In the following exercise to organize some of these ``thirty two marks", we start by assuming grammatical equivalences of the sort we have postulated elsewhere, leading to a ``periodic table" of grammatical objects. For our purposes now: 

\emph{Postulated Grammatical Equivalences (for elements relevant to Table \ref{Set1}):}  
\begin{table}[!hp]
\centering
\begin{tabular}{c c c c } 
$\pm \mathbb{I} = AP$, & $\pm X = NumP$, & $\pm Y = PredP$, &$\pm Z = NP$, \\Ê
$\pm C_1 = PP$, & $\pm C2 = VP$, & $\pm i \mathbb{I} = Elided [-1]$, & $\pm iX = IP$, \\Ê
$\pm iY = DegP$, & $\pm iZ = Elided[1]$, & $\pm iC_1 = Elided[i]$, & $\pm i C_2 = Elided[-i]$
\end{tabular}
\end{table}Ê
\vspace{-2.5em}
\subsection{Passive constructions and A-movement} 
\label{sec131}
  
Consider, next, the standard passive context in Sen.\ref{pass} (``John was arrested"), which we may suppose holds (in some combination) of A-movement as in:   
\beq
 [_{IP} ~{\rm John}_i ~ Infl-Tense ~ [_{vP} ~{\rm was} ~ [_{VP}Ê~ {\rm arrested} ~ t_i ~]_{VP} ~ ]_{vP} ~ ]_{IP}  
\label{pass}Ê
\eeq
Right away we need to worry about the trace in complement (not specifier) position.

We may assume that, while in complement position, \emph{John} becomes the specifier of \emph{arrested} by a tensor product --- then the lower trace is (vacuously, in phrasal terms) a specifier. This poses a ``timing issue". Does reinterpreting a complement as a specifier entail modifying the syntactic structure in place \emph{prior to that reinterpretation}? If so, wouldn't we have to ``overwrite" the entire phrase-marker to accommodate things? Now, in our formalism the vector representing a chain does not contain information of what happens \emph{first}, Êsince the sum of vectors is commutative, i.e., $K + \Lambda = \Lambda + K$ irrespectively of which of the two contexts $K$ or $\Lambda$ happens first in the derivation, \color{black} a fact that remains true even after we compress the chain. In other words, the compression affects both ``derivational steps" at once, in the sense that they are considered together as a superposition of vectors in the Hilbert space. Then again, is the process of reinterpreting a complement $\alpha$ (introduced via matrix multiplication with some $\Lambda$, so $\alpha \Lambda$) as a specifier $\alpha$ (yielding a tensor product $\alpha \otimes \Lambda$) in any sense comparable \emph{for the computation at large} to what happens in chain formation? If all our operations are taken to be ``simultaneous" so long as their algebraic result is identical, we have enough leeway for that\footnote{In the syntactic literature some of these processes go by the name of ``computational backtracking". While that term betrays a directional metaphor, in our system there are situations in which the way to understand what actually happened is by considering all ways it could happen and settling on one that is optimal \emph{a posteriori}, in some definable fashion. This bears also on whether a derivation is Turing-computable in standard terms (with steps coming out in sequence in the most natural implementation) or, instead, simultaneous operations obtain. Chomsky has argued for the latter in 2004, 2008, for operations within a phase.}.  

Note, next, that category $\pm i X$ is what is being postulated as $IP$ (the sentential projection). Having assumed that the ``upper context" is $\pm iX = IP$. We know that for our operations to work within a chain as presented in Table \ref{Set1} with the ``32 marks", the ``lower context" must be the orthogonal $\pm i \mathbb{I}$:  
\beqa
\det = i:&& \pm iX + i \mathbb{I}, \nonumber \\Ê
\det = -i:&& \pm iX - i \mathbb{I}.
\label{instan}
\eeqa

Note that the label in each instance above is $i$ vs. $-i$ for the chain contexts --- which affects the chain label (flipping its value). Observe, also, how $\pm i X$ relates to $\pm i \mathbb{I}$ only, which lacks a unified characteristic polynomial\footnote{ ÊThe characteristic polynomial of a matrix $A$ is the determinant of $A - \lambda {\mathbb I}$, which is a polynomial in the variable $\lambda$. We say that two matrices $A$ and $B$ have a ``unified" characteristic polynomial, if such a polynomial is the same for both matrices. \color{black}} for its ``twin" versions. In general, when ``twin" matrices (presenting the same characteristics with opposite values) have the very same characteristic polynomial, we take the grammatical system to be able to operate with such pairs. However, when the ``twin" elements lack a unified characteristic polynomial for both versions of the same grammatical category, we take the computational system not to be able to operate with such objects in relations of grammar. The question, then, is what sense it makes for $\pm i \mathbb{I}$ to be \emph{the lower context of a chain whose top anchor is $\pm iX = IP$}.

That ``instability" of the resulting chain label suggests that we cannot treat the two instances in Eqs.(\ref{instan}) on a par. At the same time we know there are two classes of chains anchored on $IP$: A-chains as in Sen.\ref{pass}, but also (more controversially for $IP$) A'-chains:
\beqa
&&{\rm John}_i {\rm, they ~ have ~ arrested}Ê~t_i. \nonumber \\Ê 
 &&[_{IP} ~{\rm John}_i ~ [_{IP} ~{\rm they} ~ Infl-Tense ~ [_{vP}Ê~ {\rm have} ~ [_{VP} ~ {\rm arrested} ~t_i ~]_{VP} ~ ]_{vP} ~ ]_{IP} ~ ]_{IP}  
\label{passAprim}Ê
\eeqa
We postulate, per the situation in Eq.(\ref{instan}), that while \emph{superposed contexts in A-chains are additive, in A'-chains they are subtractive} instead --- hence non-commutative. One immediate consequence of this postulate is that the ``parallel computations" for A'-chains may be considerably more limited, if at all present. Given that there are no present definitions in the recent literature of the A/A' chain divide, this seems reasonable.

The lack of a unified characteristic polynomial for ``twin" $\pm i \mathbb{I}$ variants entails that the grammar can only host $\pm i \mathbb{I}$ as an inert projection \emph{of the sort arising in phase transfer}\footnote{Chomsky \cite{ChomDer} argued that the computation is divided into local domains of operation (called derivational phases), which characteristically get transferred to interpretive interface domains at some critical juncture.}. This is promising in that, of course, we want lower contexts of chains to transfer to interpretation. One thing that isn't obvious about the lower context --- vis-\`a-vis the upper one anchoring the chain --- is that the latter is a ``landing site" for movements of all sorts, but surely many situations can count as ``launching sites". Witness: 
\beqa
\label{passivessen}Ê
&a.& {\rm John}_i ~{\rm was} ~  [_{VP} ~ {\rm arrested} ~t_i ~]_{VP} \\Ê
&b.& {\rm John}_i ~{\rm seems} ~  [_{AP} ~ t_i ~ {\rm angry} ~]_{AP} \nonumber \\Ê
&c.& {\rm John}_i ~{\rm has} ~  [_{NP} ~ t_i ~ {\rm a ~ temper} ~]_{NP} \nonumber \\Ê
&d.& {\rm John}_i ~{\rm was ~ seen} ~  [_{PP} ~ t_i ~{\rm with ~a ~Êgun} ~]_{PP} \nonumber \\Ê
&e.& {\rm They}_i ~{\rm seem} ~  [_{IP} ~ t_i ~ {\rm to ~have ~Êarrested ~ John} ~]_{IP} \nonumber
\eeqa
How do we secure these various (bracketed) contexts for the lower occurrence of these chains? The chain label is the product of whatever the relevant specifier is times the sum-of-contexts' label squared and normalized. The latter is fixed, if it has to be one of the orthogonal combinations in Eq.(\ref{instan}) --- and the specifier can also be fixed for all relevant examples (e.g. an NP in Sen.\ref{passivessen}). So the puzzle is:  all the lower contexts in Sen.\ref{passivessen}, despite their being $VP$, $AP$, $NP$, $PP$ or even $IP$, will have to somehow \emph{become} $\pm i \mathbb{I}$ for the system to work in these terms.
\vspace{-1em}

\subsection{Agree}
One further circumstance that we know is present in the formation of chains, as a prerequisite if Chomsky \cite{C2000} is correct, is so-called ``Agree", understood as a Probe-Goal dependency. We cannot recapitulate that in any detail here, beyond the following  summary presentation with the bare essentials (for A-chains):
\begin{enumerate} 
\item{The Probe is a (characteristic) head $H$.}
\item{The Goal is a case phrase $KP$ within the complement domain of the Probe.}
\item{The Probe has a designated, unvalued, feature $F$ that searches for an equivalent valued feature $\pm F'$ within the Goal.} 
\item{Once $\pm F'$ is encountered, $\pm F$ relates to $F$ valuating (providing a value) for $F$.}
\item{Concomitantly, the $KP$ containing $\pm F$ is turned into a regular $NP$ (without Case), at which point $NP$ becomes inactive for A-movement, except:}
\item{$NP$ may displace to the specifier of Probe $H$.}
\end{enumerate} 

From the points above, (1) and (2) are uncontroversial and have the consequence of \emph{introducing c-command and locality}, when coupled with (3)\footnote{C-command is the structural syntactic notion par excellence: (i) $X$ \emph{c-commands} $Y$ iff all categories dominating $X$ dominate $Y$, where $X \neq Y$.  There is virtually no long-range correlation in syntax that does not involve (i). The Agree relation entails c-command because the complement-domain of the Probe is its sister in the phrase-marker (2). Locality is a more nuanced notion that we will not be able to review in any detail here. Basically, though: (ii) $Y$ \emph{is local to} $X$ iff there is no $Z$ (of the same type as $X$ and $Y$) such that $Z$ c-commands $Y$ and $Z$ does not c-command $X$. One typical way in which (ii) is implemented is via the search mechanism (3).}. The controversial part of (3) pertains to the details of the valuation, more so than the specifics of the search. It is generally assumed that something of the sort in (4) happens in the Probe (the \emph{raison dÕ$\hat{{ e}}$tre} behind the Probe seeking the Goal), but it is still being discussed how. Phenomenology-wise, (5) is clear for regular A-chains, although the details of how Case is supposed to work are not; similarly, the ``deactivation" is still controversial. Finally (6) is innocuous, although it is unclear why in some languages movement is obligatory. 
 
Hardly any generative linguist (certainly few minimalists) would be too worried about us invoking something like Agree --- they may if we do not. Immediate advantages of restrictions along these lines have to do with the (a)symmetries and timing that Agree introduces, thus for instance allowing us to build chains only under c-command conditions, or only under local search conditions. Since, otherwise, chains as we have defined them could happen anywhere in a derivational space, that is a good consequence. Unfortunately, though, the specifics of Agree are unsettled enough in the literature that we will have to make our own decisions, seeing how different assumptions fit our purposes. For example, is the Agree relation a linear transformation within our matrices?

One thing is clear: if the agreement operation understood as a linear transformation were to target contexts as in the Agree conditions states above, the result couldn't possibly be a unique category like $\pm i \mathbb{I}$. This is because the contexts in point are \emph{obviously different}. At the same time, if the operation under discussion were to target \emph{the specifier} of the contexts in those conditions, then there is some hope for a unified answer, if chain specifiers themselves are unique (which seems to be generally the case, empirically: specifiers are nominal projections, modulo the issue of Case that may make them seem adpositional; they are not $VP$s or $AP$s). And recall that, per our characterizations, the anti-Hermitian ``32 marks" are all \emph{harmonic}, in the sense that the label of the chain is correlated with the label of the specifier, no matter what this specifier happens to be.

So can a specifier of the right (harmonic) sort, in combination (of some kind) with some Probe, result in a category $\pm i \mathbb{I}$ that constitutes a valid (transferred) phase in a derivation? We can observe what happens, next, when we multiply the hypothesized $Infl$ head $-iC_2$ times $\pm C_1$ (the most natural representation for a Case Phrase $KP$). There is no necessary reason why Agree in our terms should boil down to something like ``long-distance" 1st M, but that is certainly the first thing to try, given that 1st M is the one relation that we know can happen between two heads (in self-merge), and the categorial relation between the Probe and the Goal is one between two different heads:
\beq
\begin{array}{cccc}
-iC_2 = Infl &-C_1 = PP & &i\mathbb{I} = Elided[-1] \\Ê
\begin{pmatrix}
-i & 0 \\Ê
0 & 1  
\end{pmatrix} &  
\begin{pmatrix}
-1 & 0 \\Ê
0 & i  
\end{pmatrix} & =& 
\begin{pmatrix}
i & 0 \\Ê
0 & i  
\end{pmatrix}  \\
& & & \\Ê
-iC_2 = Infl &C_1 = PP & &-i\mathbb{I} = Elided[-1] \\Ê
\begin{pmatrix}
-i & 0 \\Ê
0 & 1  
\end{pmatrix} &  
\begin{pmatrix}
1 & 0 \\Ê
0 & -i  
\end{pmatrix} & =& 
\begin{pmatrix}
-i & 0 \\Ê
0 & -i  
\end{pmatrix}  
\end{array} 
\label{mats}
\eeq
The good news is that we do obtain $i\mathbb{I}$ then, which we are dubbing ``$Elided [-1]$", in order to signal the fact that it is an element that cannot be operated with further. Note that this result is irrespective of the context as in Sen.\ref{passivessen}, so long as the Probe is $-iC_2 = Infl$ (what we need to assume in order to say that a category 1st merging with $vP$ yields $\pm i X = IP$ as its projection) and the specifier this Probe relates to is a $\pm iC_2 = PP = KP$. 

Note, also, how the operation in Eq.(\ref{mats}) makes the Agree essentials (2)/(5) necessary too, for no other Agree understood as long-distance 1st M of $-iC_2 = Infl$, with any of the other objects in $G_{pc}$, will produce the desired result. Since presently it is not understood why the grammar manifests the phenomenon of Case to begin with, this is not insignificant. That approach leaves us with two major concerns to deal with. 

\subsection{Consequences for compositionality and case} 

The first concern should be what any of this may mean semantically, and specifically whether it destroys compositionality. If semantics is computed in Montogovian terms (for which the meaning of whatever the derivation constructs is assumed to be computed as a function of the meaning of the parts and how they combine), it is problematic to say that a $VP$, $AP$, $NP$, $PP$, or $IP$ are seen as some element, \emph{the same element}, with label $[-1]$ (or any other). Then again, the Montogovian view is a hypothesis, and not the only hypothesis about how meaning is computed; matters are subtler in Neo-Davidsonian terms. In the latter, all sentences are existential event quantifications, with a main predicate, key arguments, and indefinitely many modifiers, roughly as in the following:
\beqa
&a.& \exists x ~ CAUSE(x) ~ [ \exists y ~ {\rm arrest} (y) ~[ AGENT ~({\rm someone},x) ~THEME ~({\rm John}, y) ]] \nonumber \\Ê
&b.& \exists x ~ EXPERIENCE(x) ~ [ \exists y ~ SENSATION (y) ~[ PATIENT ~({\rm John},x) ~THEME ~({\rm anger}, y) ]] \nonumber \\Ê
&c.& \exists x ~ temper(x) ~ [ INTEGRAL ~ PREDICATION ~ ({\rm John},x) ] 
\label{sen12}Ê
\eeqa
In these kinds of representations, syntactic brackets are only of moderate importance to semantic representation, beyond keeping track of dependencies: what is crucial is to establish the right scope relations between the various event quantifiers, to determine the specifics of the thematic roles employed in each instance ($THEME, ~ PATIENT, ~ AGENT$) and to fix relevant event predicates (modifiers) that may be conjunctively added to a given event variable. All of that is certainly recoverable even if the syntactic process obliterates information about $VP$, $AP$, $NP$, and the like, as such\footnote{That sort of information is semantically superfluous --- we need to get rid of it if the systemÕs primitives are predicates, arguments, roles, quantifiers, or variables. From that perspective, just as we want the level of PF to eventually eliminate phrasal brackets that help us build rhythmic patterns and intonation contours, so too we want the level of LF to brush off phrasal labels, once we have established semantic relations.}. 

In addition to that semantic consequence, our system faces a syntactic issue, which is in the details of forming a chain between the lower and upper contexts, as in Eq.(\ref{archain}). Whereas it is easy to see how to factor out $\alpha$ in the left side of that equation (leading to the right side of the equation), it is harder to see how to proceed in an instance in which, unless something is remedied, what we have in the lower context (at least prior to agree) is a $KP$ which we are hypothesizing is $\pm C_1$, whereas what we have in the upper, anchoring, context (after agree) is, rather, a $NumP$ or, at any rate, some nominal element with label $-1$ --- with the Case gone: so not possibly $\pm C_1$ with label $-i...$

This must mean that, although the $\pm C_1 = KP$ agreeing with $-iC_2 = Infl$ transfers as some $[-1]$ inert $\pm i \mathbb{I}$ element, which will constitute ``the lower context of the chain", the specifier of this chain cannot be the $\pm C_1 = KP$ that led precisely to that transfer, but must instead be \emph{the very same element that displaces} to $\pm i \mathbb{I} = IP$. In other words, as we construct the chain, we must assume that what enters into a tensor product with the transferred $\pm i \mathbb{I}$ is \emph{the substantive part of the} $\pm C_1 = KP$. Syntactically this makes sense. When we operate with \emph{John's}, including the genitive, we of course know that the case marker is just a feature; substantively the ``real thing" is a nominal projection \footnote{In fact, the Jarret graph itself prevents us from first merging a $KP$ in relevant domains (as a complement of $V$ or $P$); this must mean either that case features are ignored for Merge altogether, or that they somehow emerge as a consequence of Agree.}. However, we now have a precise system, and thus for instance the Jarret graph treats differently an element of the adpositional sort than it does an element of the nominal sort. So we will need a further operation that \emph{forms the chain with the right element}, after transfer.

There is no contradiction in that claim. We have said that $\pm i \mathbb{I}$ is semiotically not operable with because it lacks a unified characteristic polynomial for its ÒtwinÓ variants --- this is behind the idea of transfer, since these elements are derivationally useless, by hypothesis. Of course, the operation of forming the relevant chain, as such, is terminal: after we do the superpositions, for which we do need the relevant tensor products that $\pm i \mathbb{I}$ has to participate on, there won't be any further use of said contexts. There are only two fates awaiting them. If the chain is not observed at that phase, its mathematical workings will continue internal to the derivation, but their own specific significance will be lost. If, on the other hand, the chain is observed at that phase, chances are that the top or bottom context, but not both, will be interpreted. And there is, then, the chance that the $\pm i \mathbb{I}$ category has to be visible to the system, \emph{at the juncture it is observed}.

Another way of finessing the point is that the issue with unified characteristic polynomials for ``twin" categories is problematic only \emph{for an indefinite long-range correlation determined by that category}, since the system needs to know that it is dealing with a category of some sort --- the ``twin" variants, as such, being irrelevant. But in situations of locality there should be no issue of this sort. Locality here is relative, in that several processes are happening at once within the phase. In particular, the rationale behind this idea is that a chain starts with a $KP$ specifier, which gets to be probed into, which turns it into an elided category of a neutralized (nominal) sort --- a valid lower context for a chain. But the chain is not formed with that $\pm C_1 = KP$, but instead \emph{with the substantive nomimal that the $\pm C_1 = KP$ contains}. Logically speaking, the chain has a ``before" with a $\pm C_1 = KP$ and an ``after" with the substantive nominal within. Yet all of this is taken to be the optimal set of relations forming the chain: the Probe searching for the Goal, the match, the transfer of the phase material, and the formation of the chain with the valid contexts and specifiers. All at once.

Having shown one specific instance at work, we will not discuss matters further in this context. Elsewhere in our work we have studied other pairings from the ``32 marks" in Table \ref{Set1} as in Sen.\ref{pass}, extending our discussion to subtractive A'-chain conditions and beyond. In fact, at the beginning of this section we alluded to constructions beyond standard chains (predications, head-to-head dependencies), which our system allows us to treat as well, but which would be unreasonable for us to present now --- they create analytical issues similar to the ones discussed above for chains anchored on $\pm i \mathbb{I} = IP$. 

\clearpage 

\section*{\huge Conclusions and some extra thoughts}Ê
\addcontentsline{toc}{section}{\protect\numberline{}Conclusions and some extra thoughts}%
\label{sec13} 

In this paper we have presented the mathematical foundations of MS, a model of syntax that describes external and internal merge relations, from the core head-complement and head-specifier local dependencies, to long-range correlations of the sort arising in chains. The model is based on a small set of axioms (which could be reduced even further), from which a large amount of structure emerges. Importantly, several groups and Hilbert spaces can be derived in a natural way from the axioms of the theory, which turn out to have a natural interpretation in terms of observed language properties. Moreover, some of the mathematical aspects of the model resemble those of quantum mechanics, which we find particularly intriguing. It is also important to remark that MS turns out to be an arguably simple theory of long-range correlations in language. As such, however, MS is still a theory under development and evolution. What has been presented here is its mathematical foundation only, and how that enables a natural explanation of the distributed nature of chain occurrences.

We understand this work as the beginning of a research program. We have chosen one particular representation stemming from Chomsky's original work, because we find that work deep, insightful and productive. That, however, does not guarantee that this representation is the actual one that, in some sense or another, human mind/brains use; indeed, we have cited other representations that also purport to be \emph{of language} and which are, nevertheless, distinct from the one chosen here. All of those cannot be correct --- perhaps none of them is. But this is the spirit in which we have written the present paper: as a proof of concept, inasmuch as we have argued that a proper treatment of long-range correlations in language ought to be \emph{of this sort}. Indeed, even within our particular choice one could use, for instance, a different inner product, with different results, or different auxiliary assumptions, and so on. 

We have tried to be relatively ``close to linguistic data", especially in distributional terms (local dependencies or chain conditions that do or do not arise across languages, according to data reported in the standard linguistic data). There are, however, scores of situations carefully discussed in the literature that we did not address in any detail here. For instance: what about more complex types of chains and possible entanglements in other types of derivations? That topic is fair game for a project of this sort, and success or failure in the particular choice of theory we have made ought to be measured, first, against progress towards understanding such situations. In addition, familiar questions for any such program remain: Is there a simpler mathematical formalism to account for the structure presented here? Are there mathematical symmetries in the model accounting for known linguistics restrictions? Did the mathematical similitude with quantum mechanics happen just by chance, or does it have some deeper meaning? 

We want to end by considering a few extra thoughts, both technical and conceptual in nature. We start with the so-called Òderivational dualityÓ: some processes proceed derivationally, in some fashion, while the results of others are ``arguably stored (for a while)". In a sense, this is not surprising, for automata traditionally postulated to recognize minimally complex formal languages within the Chomsky Hierarchy typically require a ``stack" or some such memory device. At the same time, our approach suggests that the role of what goes into that memory device --- which may need to be more nuanced than a ``stack"  ---  allows the system to run alternative derivations, selecting an optimal path a posteriori. This reminds us of superpositions in vector spaces, suggesting that the ultimate edifice of the language faculty is a bit more elaborate than the standard classical Turing architecture would lead us to expect. We will not attempt to propose an alternative computational apparatus, although we do know that quantum computation can clearly do this and more; indeed, the issue is the ``and more" bit. Whatever computation architecture may ultimately be needed to implement our suggestions obviously needs to be restricted so as not to permit just any imaginable computation, for language processes are as restricted as they are nimble.

That may ultimately be the biggest difference between our approach and others in the literature implementing neural networks in terms of a Hilbert space (in some variation). We are interested in modelling Minimalist Syntax, not questioning it. But we do take very seriously the idea that, in such a syntax, a limited set of derivational ``parallel tracks" may be needed. We have shown our reasons, which can be evaluated or criticized, seeking alternatives if those arise and are preferable. We have also tried to ponder the limitations that our system imposes. For example, we have emphasized the asymmetry of A vs. A'-relations, where we actually do not expect the same sorts of parallel computations, and we do, instead, field effects, which should reduce computational load: by preventing some derivations (violating locality). In the sort of architecture we have in mind, the computational implementation would not even try derivations alternative to the ones meeting the field structures.

All of that possibly relates to what ``stabilizes into a lexicon", which once fixed allows speakers to construct derivations from its useful atoms. In particular, while the ``computational buffer" may allow certain elements to have a ``virtual" existence in a derivation in process --- until their interpretive fate is resolved upon the collapse and externalization of their derivational vector --- such ``ghosts" may turn real if frequent and meaningful enough, in the form of atomized frozen idioms or eventually words. Presumably, then, said structures would move from the virtual status of short-term derivational (procedural?) memory to the long-term lexical (declarative?) memory that allows instant recall at a later use, as lexical atoms.

Seen this way, it may even be that the entire edifice of syntax is a \emph{plastic memory device} that effectively builds a culture in the form of a lexicon: by way of ``ghost thoughts" that emerge in the interaction with human conspecifics and the environment. Surely that allows for ``communication" (or, also, lying, thinking, emoting, etc.); but physiologically it would be literally constructing and reinforcing a lexicon \emph{as it acts}. Then acquiring a language during the critical, most plastic, years would not be so much ``learning" it as literally creating a robust lexicon from what the device allows as permissible interactions. If correct, this view purges the system from a variety of functionalist residues. For example, we shouldn't really need ``phrasal axioms" as such to map a system like ours to the semantics, any more than we need them to map it to the phonetics: whatever is possible should be so because it is natural, in a sense to be made specific as understanding advances. We haven't said much about the phonetic mapping, but we have offered some ideas about the semantic one, which piggy-backs on such formal niceties as expected (a)symmetries or whether our operators are hermitian. 

Ultimately from this perspective, the issue is not so much: ``Do we need (say) phases to reduce computational complexity when communicating a thought to others?" --- but rather: ``What is the optimal way in which a system of these characteristics stabilizes as a lexicon?" That is meant quite literally, in the sense one can ask what the optimal conditions in the universe are that allow stabilization into what we observe. For example, it may be that the ``size" of a phase ultimately has to do with the size of what can lexicalize in the shape of an idiom. If that were true, there would be no phases larger than, well, the shape of the largest idiom. Evidently, that would only push the agenda to the next set of questions: Why are those the right sizes, etc. But that is every bit as testable as everything else we find in the realm of cosmology, particle physics, chemistry, physiology, and so on.

To conclude: the mathematical foundation of MS presented in this paper is only the tip of some iceberg. It is now time to dig deeper into its mathematical conditions, match that with the known linguistic reality, and understand if there are more basic principles at work in human language. The future looks busy and, we feel, rather entertaining. It remains to be seen whether it is actually meaningfully real too.

\clearpage 

\noindent {{\bf Acknowledgements:}} we acknowledge fruitful discussions with B. Berwick, C. Boeckx, J. Colarusso, A. Gallego, B. Idsardi, M. Jarret, D. Krivochen, H. Lasnik, D. Medeiros, M. Piattelli-Palmarini, D. Saddy, Z. Stone, and P. Vitiello on several aspects of this project. We also acknowledge the \emph{Universitat Aut\`onoma de Barcelona}, where many key discussions concerning this work took place.

\end{document}